# Florentin Smarandache

# INTRODUCTION TO NEUTROSOPHIC STATISTICS

*2014*

# Contents

# Introduction

Although the neutrosophic statistics has been defined since 1996, and published in the 1998 book **Neutrosophy. / Neutrosophic Probability, Set, and Logic**, it has not been developed since now. A similar fate had the neutrosophic probability that, except a few sporadic articles published in the meantime, it was barely developed in the 2013 book "Introduction to Neutrosophic Measure, Neutrosophic Integral, and Neutrosophic Probability".

Neutrosophic Statistics is an extension of the classical statistics,and one deals with set values instead of crisp values.

In most of the classical statistics equations and formulas, one simply replaces several numbers by sets. And consequently, instead of operations with numbers, one uses operations with sets. One normally replaces the parameters that are indeterminate (imprecise, unsure, and even completely unknown). That's why we made the convention that any number $a$ that is replaced by a set be noted $a_N$, meaning *neutrosophic a*, or *imprecise, indeterminate a*. $a_N$ can be a neighbour of $a$, can be an interval that includes $a$, and in general it can be any set that approximates $a$. In the worst scenario, $a_N$ can be unknown. In the best scenario (when there is not indeterminacy related to $a$), $a_N = a$.

Why this passage from crisp numbers to sets? Because in our real life we cannot always compute or provide exact values to the statistics characteristics, but we need to approximate them. This is one way to passing from classical to neutrosophic statistics, but other ways could be possible, depending on the types of indeterminacies, and the reader is kindly invited to do such research to be published in the next issues of the international journal of "Neutrosophic Sets and Systems", http://fs.gallup.unm.edu/NSS/.

The author would like to thank Prof. Yoshio Hada, the President of Okayama University of Science, Prof. Valery Kroumov from Okayama University of Science, Prof. Akira Inoue from the State University of Okayama, also Prof. Masahiro Inuiguchi, Dr.Masayo Tsurumi, and Dr. Yoshifumi Kusuroku from the University of Osaka, and Dr. Tomoe Entani from the Hyogo University for their valuable considerations and opinion during my postdoctoral research in Japan in December 2013 and January 2014 about applications of the neutrosophic science in robotics and other fields.

Any quantity computed with some indeterminacy from values in a sample (i.e. not exactly) is a neutrosophic statistics.

A neutrosophic statistic is a random variable and as such has a neutrosophic probability distribution.

The long-run behaviour of a neutrosophic statistic's values is described when one computes this statistic for many different samples, each of the same size.

Neutrosophic Statistics is an extension of the classical statistics. While in classical statistics the data is known, formed by crisp numbers, in neutrosophic statistics the data has some indeterminacy.

In the neutrosophic statistics, the data may be ambiguous, vague, imprecise, incomplete, even unknown. Instead of crisp numbers used in classical statistics, one uses sets (that respectively approximate these crisp numbers) in neutrosophic statistics.

Also, in neutrosophic statistics the sample size may not be exactly known (for example the sample size could be between 90 and 100; this may happen because, for example, the statistician is not sure about 10 sample individuals if they belong or not to the population of interest; or because the 10 sample individuals only partially belong to the population of interest, while partially they don't belong).

In this example, the neutrosophic sample size is taken as an interval n = [90, 100], instead of a crisp number n = 90 (or n = 100) as in classical statistics.

Another approach would be to only partially consider the data provided by the 10 sample individuals whose membership to the population of interest is only partial.

# Neutrosophic Statistics

Neutrosophic Statistics refers to a set of data, such that the data or a part of it are indeterminate in some degree, and to methods used to analyze the data.

In Classical Statistics all data are determined; this is the distinction between neutrosophic statistics and classical statistics.

In many cases, when indeterminacy is zero, neutrosophic statistics coincides with classical statistics.

We can use the neutrosophic measure for measuring the indeterminate data.

The neutrosophic statistical methods will enable us to interpret and organize the neutrosophic data (data that may have some indeterminacies) in order to reveal underlying patterns.

There are many approaches that can be used in neutrosophic statistics. We present several of them through examples, and afterwards generalizations for classes of examples. Yet, the reader can invent new approaches as well in studying the neutrosophic statistics.

We emphasize, as in neutrosophic probability, that **indeterminacy** is different from **randomness**. While classical statistics is referring to randomness only, neutrosophic statistics is referring to both randomness and especially indeterminacy.

**Neutrosophic Descriptive Statistics** is comprised of all techniques to summarize and describe the neutrosophic numerical data's characteristics.

Since neutrosophic numerical data contain indeterminacies, the neutrosofic line graphs, and neutrosophic histograms are represented in 3D-

spaces, instead of 2D-spaces as in classical statistics. The third dimension, in addition of the XOY Cartesian System, is that of indeterminacy (I). From unclear graphical data displays we can extract neutrosophic (unclear) information.

**Neustrosophic Inferential Statistics** consists of methods that permit the generalization from a neutrosophic sampling to a population from which it was selected the sample.

**Neutrosophic Data** is the data that contains some indeterminacy.

Similarly to the classical statistics it can be classified as:

- **discrete neutrosophic data**, if the values are isolated points; for example: $6 + i_1$, where $i_1 \in [0, 1], 7, 26 + i_2$, where $i_2 \in [3, 5]$;

- and **continuous neutrosophic data**, if the values form one or more intervals, for example: $[0, 0.8]$ or $[0.1, 1.0]$ (i.e. not sure which one).

Another classification:

- **quantitative** (numerical) **neutrosophic data**; for example: a number in the interval [2, 5] (we do not know exactly), 47, 52, 67 or 69 (we do not know exactly);

- and **qualitative** (categorical) **neutrosophic data**; for example: blue or red (we don't know exactly), white, black or green or yellow (not knowing exactly).

Also, we may have:

- **univariate neutrosophic data**, i.e. neutrosophic data that consists of observations on a neutrosophic single attribute;

- and **multivariable neutrosophic data**, i.e. neutrosophic data that consists of observations on two or more attributes.

As a particular cases we mention the **bivariate neutrosophic data**, and **trivariate neutrosophic data**.

A **Neutrosopical Statistical Number $N$** has the form:

$$N = d + i,$$

where $d$ is the determinate (sure) part of $N$, and $i$ is the indeterminate (unsure) part of $N$.

For example, $a = 5 + i$, where $i \in [0, 0.4]$, is equivalent to $a \in [5, 5.4]$, so for sure $a \geq 5$ (meaning that the determinate part of $a$ is 5), while the indeterminate part $i \in [0, 0.4]$ means the possibility for number „a" to be a little bigger than 5.

We may consider, similarly to the classical statistics, a **neutrosophic stem-and-leaf display** of data.

For example, let's have the neutrosophic data that follows:

$$6 + i_1, \text{with} i_1 \in (0, 0.2);$$
$$7 + i_2 \text{with} i_2 \in [2, 3];$$
$$6 + i_3, \text{with} i_3 \in [0, 1];$$
$$9 + i_4, with i_4 \in [1.1, 1.5);$$
$$9 + i_1.$$

Its neutrosophic stem-and-leaf display is:

$$
\begin{array}{c|cc}
6 & i_1 & i_3 \\
7 & & i_2 \\
9 & i_1 & i_4
\end{array}
$$

or under the form of interval:

$$
\begin{array}{c|cc}
6 & (0, 0.2) & [0, 1] \\
7 & & [2, 3] \\
9 & (0, 0.2) & [1.1, 1.5]
\end{array}
$$

Obviously a **neutrosophic statistic number** can be written in many ways.

If you retake: $a = 5 + i$, with $i \in [0, 0.4]$, then $a = 4 + i_1$, with $i \in [1, 1.4]$, or $a = 3 + i_2$, with $i_2 \in [2, 2.4]$, and in general $a = \propto + i_\propto$, with $i_\propto \in [5-\propto, 5.4-\propto]$, and $\propto$ any real number.

Or in opposite way:

$a = 5.4 - i_3$, with $i_3 \in [0, 0.4]$, and in general

$a = \beta - i_\beta$, with $i_\beta \in [\beta - 5.4, \beta - 5]$, and $\beta$ any real number.

A **Neutrosophic Frequency Distribution** is a table displaying the categories, frequencies, and relative frequencies with some indeterminacies. Most often, indeterminacies occur due to imprecise, incomplete or unknown data related to frequency. As a consequence, relative frequency becomes imprecise, incomplete, or unknown too.

An example about the neutrosophic frequency distribution concerning the number of accidents by car drivers.

| Number of accidents | Neutrosophic frequency | Neutrosophic relative frequency |
|---|---|---|
| 0 | 50 | [0.185, 0.227] |
| 1 | [60, 80] | [0.240, 0.333] |
| 2 | [70, 90] | [0.280, 0.375] |
| 3 | [40, 50] | [0.154, 0.217] |
| Total 0-3 | [220, 270] | [0.859, 1.152] |

How to read the previous table, let's say line #2: the number of car drivers with only one accident is between 60 and 80 (thus unclear information), and corresponding neutrosophic relative frequency between 0.2240 and 0.333.

To compute the total for the neutrosophic frequencies, where we have imprecise information, we compute the *min* and *max* of estimated frequencies:

$$min_{nf} = 50 + 60 + 70 + 40 = 220,$$
$$\text{and} max_{nf} = 50 + 80 + 90 + 50 = 270.$$

To compute the neutrosophic relative frequency, we also do the *min* and *max* of all possibilities.

For zero accidents:

$$min_{nrf} = \frac{50}{270} \simeq 0.185$$
$$\text{and} max_{nrf} = \frac{50}{220} \simeq 0.227,$$
$$\text{or } 50 \div [220, 270] \simeq [0.185, 0.227].$$

For one accident one has:

$$min_{nrf} = \frac{60}{50 + 60 + 90 + 50} = 0.240$$

$$max_{nrf} = \frac{80}{50 + 80 + 70 + 40} \simeq 0.333.$$

For two accidents one has:

$$min_{nrf} = \frac{70}{50 + 80 + 70 + 50} = 0.280,$$

$$\text{and} \, max_{nrf} = \frac{90}{50 + 60 + 90 + 40} \simeq 0.375.$$

The interval [0.280, 0.375] is different from:

$$[70, 90] \div [220, 270] = \left[\frac{70}{270}, \frac{90}{220}\right] \simeq [0,259, 0.409].$$

For three accidents one has:

$$min_{nrf} = \frac{40}{50 + 80 + 90 + 40} \simeq 0.154$$

$$\text{and} \, max_{nrf} = \frac{50}{50 + 60 + 70 + 50} \simeq 0.217$$

and similarly the interval [0.154, 0.217] is different from:

$$[40, 50] \div [220, 270] = \left[\frac{40}{270}, \frac{50}{220}\right] \simeq [0.148, 0.227].$$

We simply cumulated the neutrosophic relative frequencies as an addition of intervals:

$$[0.185, 0.227] + [0.240, 0.333] + [0.280, 0.375]$$
$$+ [0.154, 0.217] = [0.859, 1.152].$$

**Neutrosophic Statistical Graphs** are graphs that have indeterminate (unclear, vague, ambiguous, unknown) data or curves.

1.a. **Neutrosophic Bar Graph**:

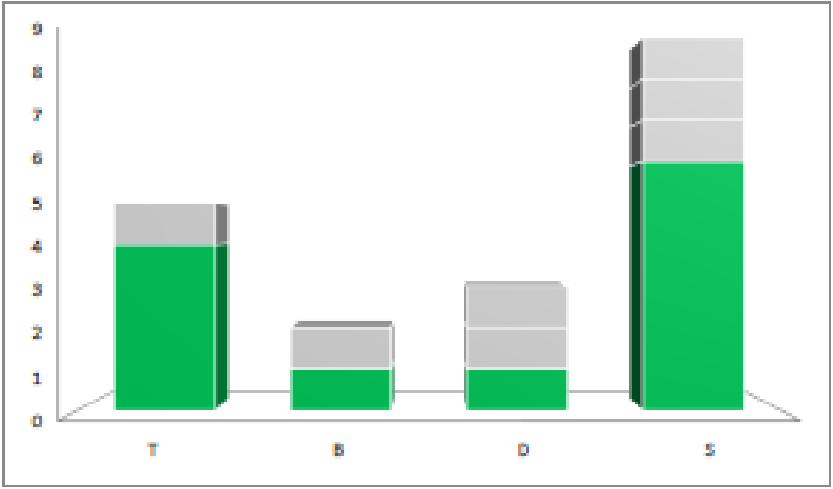

Table: *Time spent by an American daily*
T=watching TV: between [4,5] hours;
B=reading books: between [1,2] hours;
D=driving: between [1,3] hours;
S=sleeping: between [6,9] hours.

2.a. **Neutrosophic Circle Graph** for the same example:

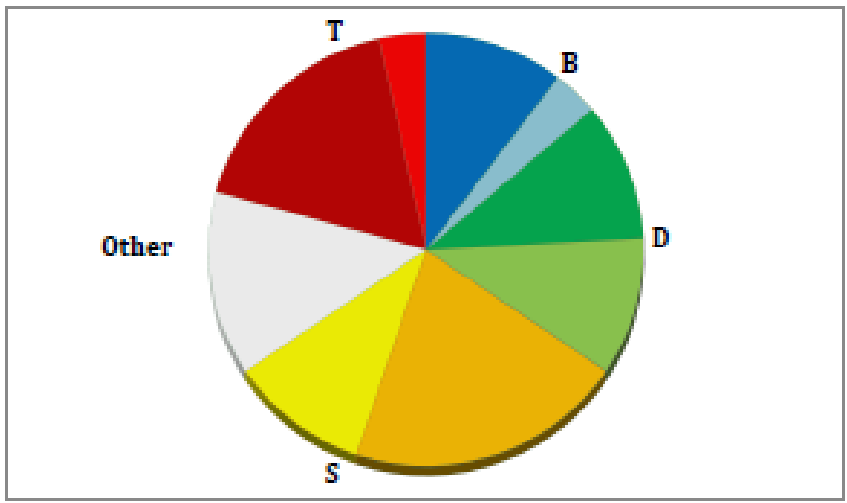

3.a. **Neutrosophic Double Line Graph** for the same example:

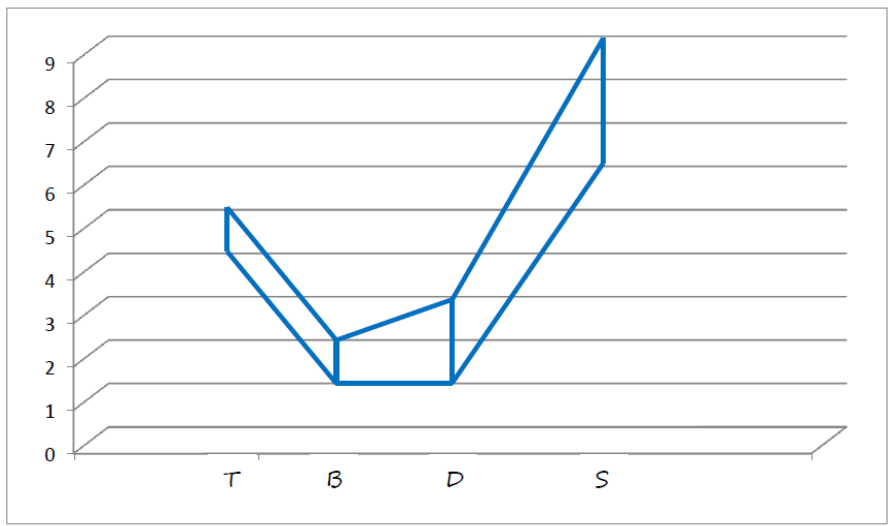

4.a. **Neutrosophic Line Plot** for the same example:

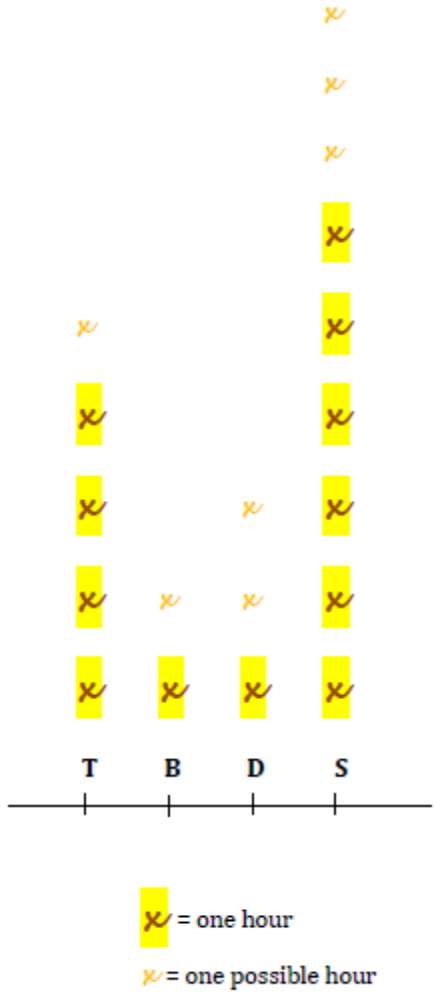

5.a. **Neutrosophic Pictograph** for the same example:

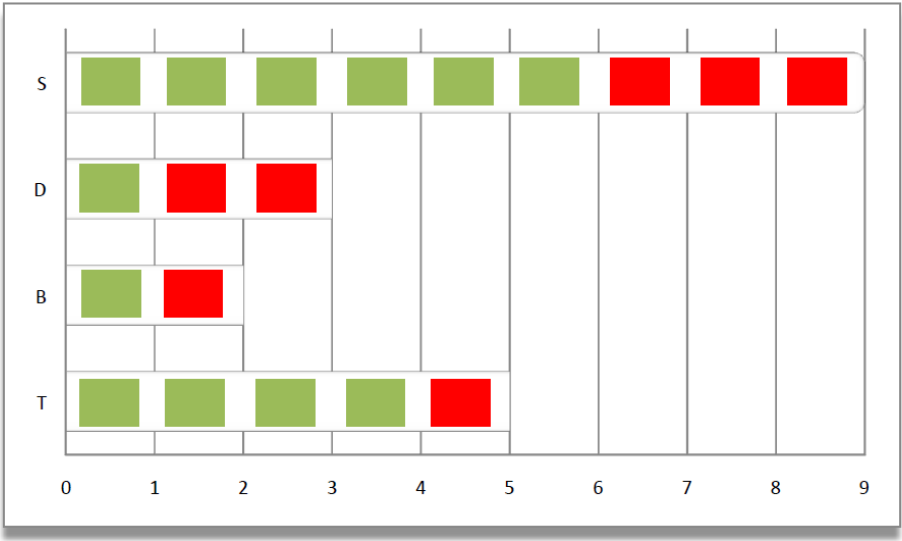

<p align="right">Green color rectangle: one hour<br>Red color rectangle: one possible hour</p>

6.a. **Neutrosophic 2D Histogram** is a neutrosophic bar graph such that the bars are vertical, there is no gap between bars (the bars of height zero are also included), and the width of each bar has the size of the represented interval. It shows, within a certain interval, the approximate number of times data occur.

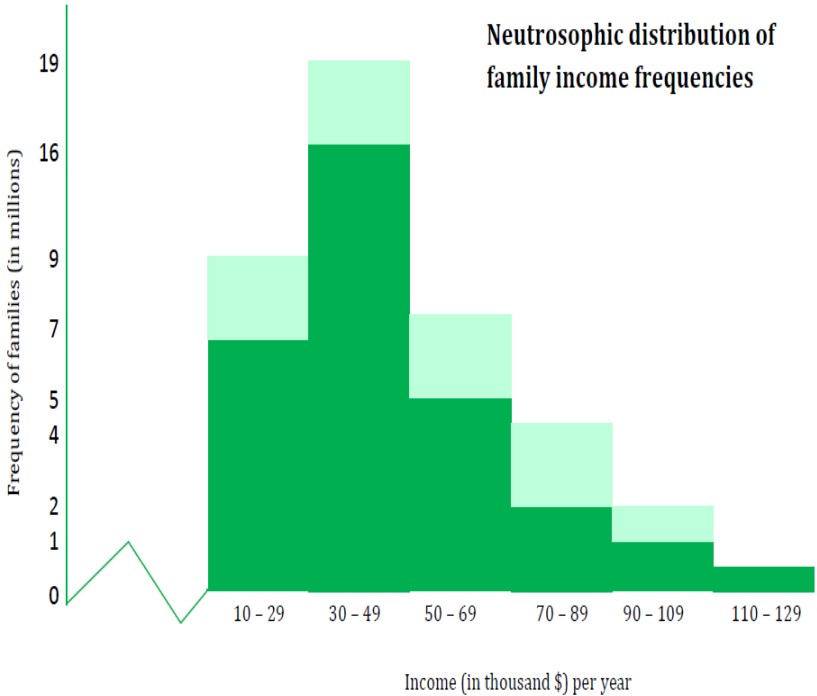

where 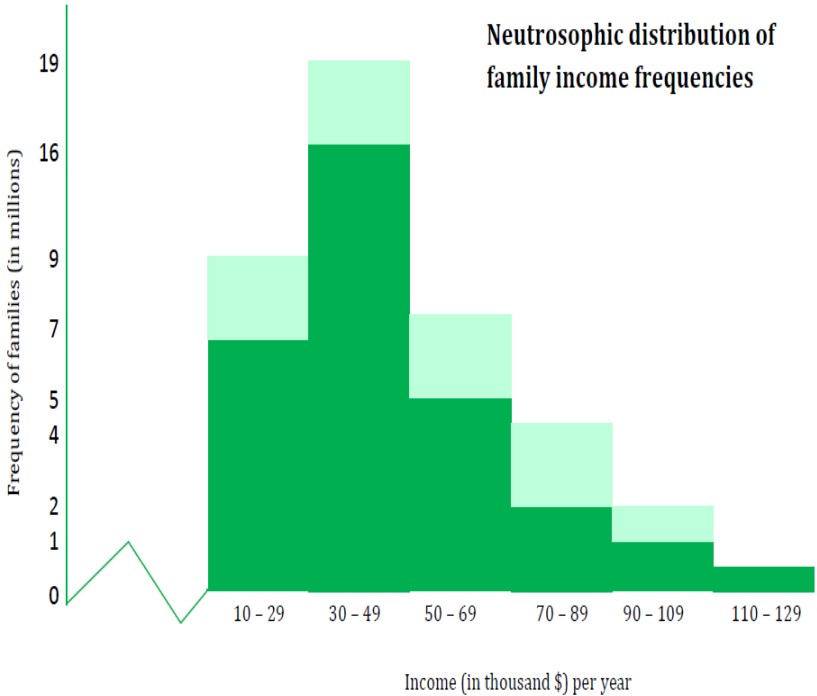 indicates in the numbering scale a distortion.

The frequencies are not crisp numbers as in classical statistics, but between some limits. For example, the number of families with income in between $10,000 – $29,000 is between 7 and 9 millions of families. Similarly for other classes of income, except for the last class of income in between $110,000 – $129,000 that corresponds to a crisp number: 1 million of families.

We represented all types of neutrosophic statistical graphs in a space of dimension two (2D) as in classical statistics, but it is also possible to make the graphs in a space o dimension three (3D), just adding to each of the previous 2D-graphs an indeterminate dimension, which measures the indeterminacy of the data.

### 1.b. **The Neutrosophic 3D Bar Graph**

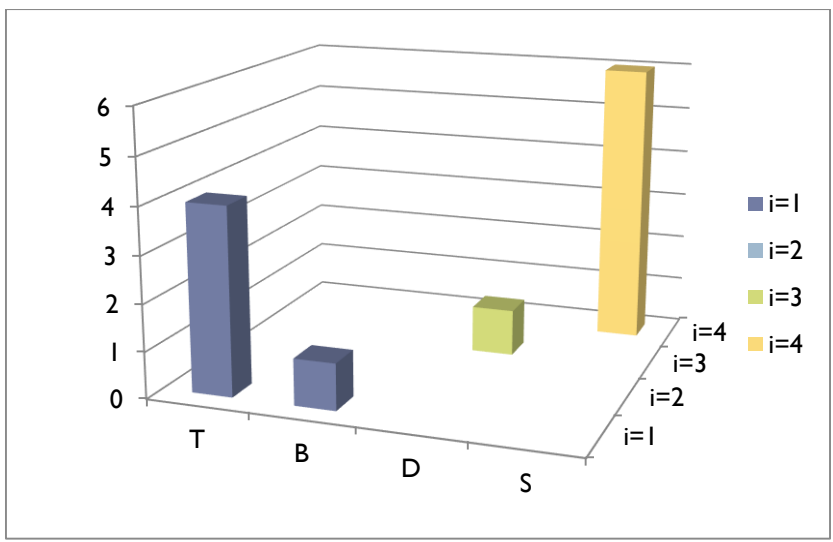

The deepness axis (i) measures the indeterminacy.
For the previous example: Time spent by an American daily.

## 2.b. **Neutrosophic Cylinder Graph**

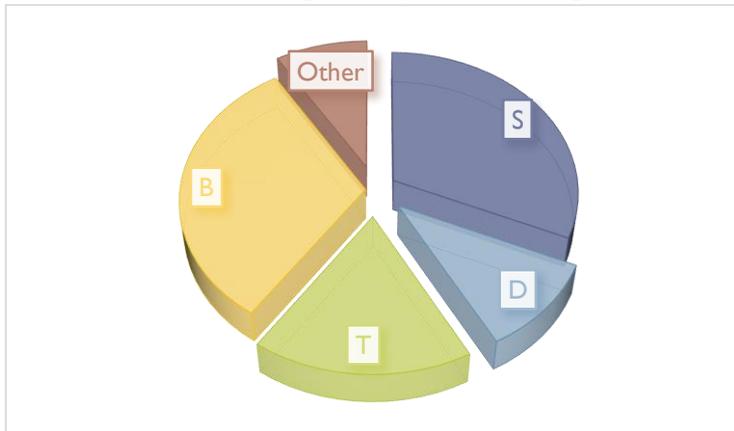

The heights (that represent indeterminacies) of T and B are the same, while the height of D is double, and the height of S is triple.

## 3.b. **TheNeutrosophic 3D-Line Graph**

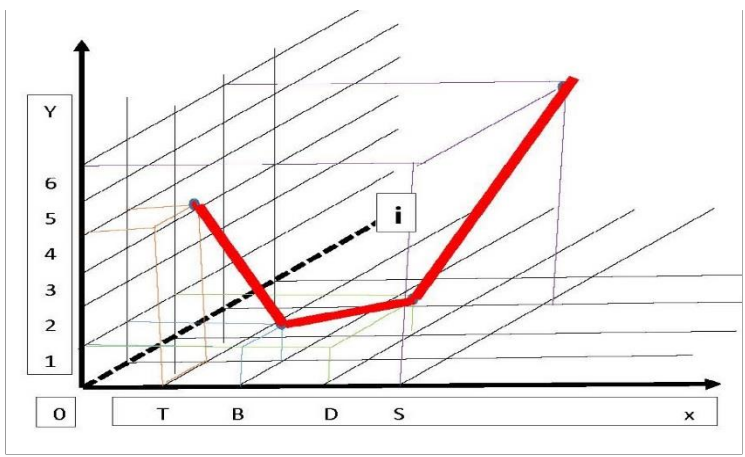

for the same example. We plot the points of coordinates (T, 4, 1), (B, 1, 1), (D, 1, 2), and (S, 6, 3), where the second component represents the determinate part ($y$) and the third component the maximum indeterminacy ($i$), and connect them. We get a 3D curve.

4.b. **Neutrosophic 3D Plot** for the same example:

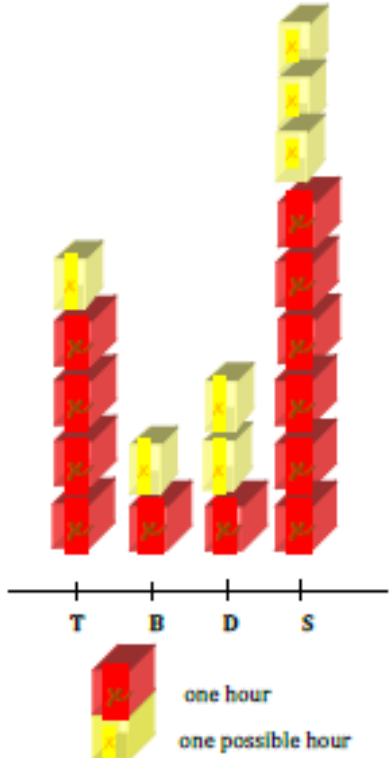

one hour

one possible hour

5.b. **Neutrosophic 3D Pictograph** for the same example:

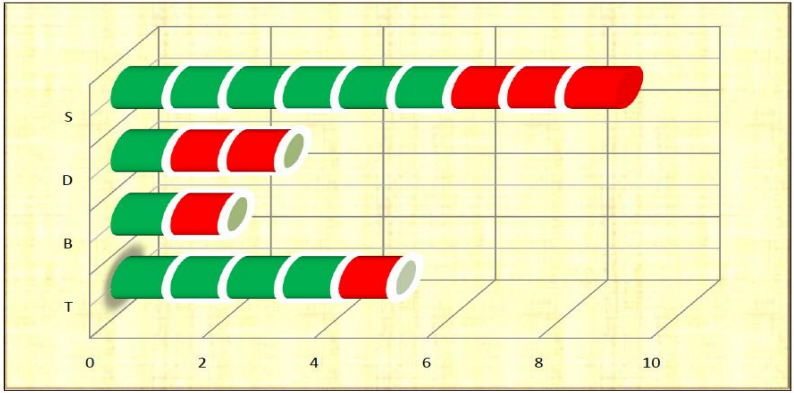

6.b. **Neutrosophic 3D Histogram** for the same example of Neutrosophic distribution of family income frequencies:

**Neutrosophic distribution of family income frequencies**

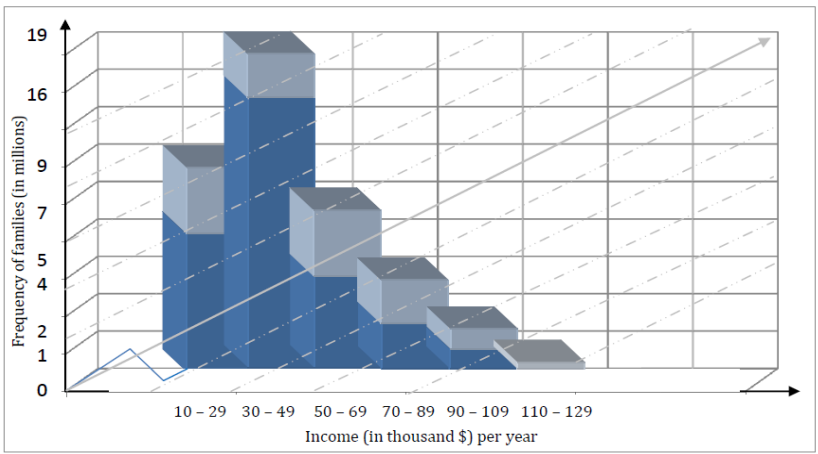

**Statistical Deceptions** can be expressed in the neutrosophic way. For example:

a) "Company's heating bill went up to 10% last year." In a neutrosophic way we can write: [0, 10]% (which could be any number between 0 and 10, including the extremes).

b) "We guarantee you lose as much as 15 pounds in a month, or your money back." Actually you lose [0, 15] pounds, so you may lose no pound!

c) "No product is better than Brian's." This means that other products could be the same as Brian's!

# Neutrosophic Quartiles

Let's consider the set of neutrosophic observations of a variable listed in almost ascending order (since we deal with sets instead of crisp numbers we have a partial order).

The neutrosophic quartiles are similarly as in classical statistics defined: the first (lower quartile) is the $\frac{1}{4}(n+1)th$, the second is the $\frac{2}{4}(n+1)th$, and the third the $\frac{3}{4}(n+1)th$.

If $(n+1)$ is not divisible by 4, then one takes the average of the two neutrosophic observations whose ranks the quartile falls in between. Another procedure is to take the inferior integer part of $\frac{i}{4}(n+1)$, for $i = 1, 2, 3$.

Let's compute the midpoint of a set ⊔ in the following way:

midpoint ⊔ $= \frac{\inf⊔ \,+\, \sup ⊔}{2}$.

We can define a *total order* on the $n$ neutrosophic observation sets in the following way:

for any sets ⊔ and ∨ we have ⊔<∨ if
$\begin{cases} \text{either midpoint}\,(⊔) < \text{midpoint}(∨), \\ \text{or midpoint}\,(⊔) = \text{midpoint}(∨) \text{and} \min ⊔ < \min ∨ \,. \end{cases}$

If it happens that

midpoint $(⊔) = $ midpoint$(∨)$

and min ⊔$=$ min ∨,

then automatically max ⊔ $=$ max∨, therefore

⊔≡∨.

An example with n = 12 ascending neutrosophic observations:

$$1, (2,3), \boxed{\{4,6\},5}, [7,10], \boxed{[7,11],9}, 12, \boxed{14, [14,15]}, 20,$$
$$\{21\} \cup (22,25].$$

First quartile:

$$\frac{1}{4}(n+1) = \frac{1}{4}(12+1) = 3.25,$$

then we average the 3rd and the 4th ranked observations:

$$\frac{\{4,6\}+5}{2} = \frac{\{4+5,6+5\}}{2} = \frac{\{9,11\}}{2} = \left\{\frac{9}{2}, \frac{11}{2}\right\} = \{4.5, 5.5\}.$$

Second quartile:

$$\frac{2}{4}(n+1) = \frac{2}{4}(12+1) = 6.50,$$

then we average the 6th and 7th ranked observations:

$$\frac{[7,11]+9}{2} = \left[\frac{7+9}{2}, \frac{11+9}{2}\right] = [8,10].$$

Third quartile:

$$\frac{3}{4}(n+1) = \frac{3}{4}(12+1) = 9.75,$$

then we average the 9th and 10th ranked observations:

$$\frac{14+[14,15]}{2} = \left[\frac{14+14}{2}, \frac{14+15}{2}\right] = [14, 14.5).$$

# Neutrosophic Sample

A Neutrosophic Sample is a chosen subset of a population, subset that contains some indeterminacy: either with respect to several of its individuals (that might not belong to the population we study, or they might only partially belong to it), or with respect to the subset as a whole. While the classical samples provide accurate information, the neutrosophic samples provide vague or incomplete information.

By language abuse one can say that any sample is a neutrosophic sample, since one may consider their determinacy equals to zero.

**Neutrosophic Survey Results** are survey results that contain some indeterminacy.

A **Neutrosophic Population** is a population not well determined at the level of membership (i.e. not sure if some individuals belong or do not belong to the population).

For example, as in the neutrosophic set, a generic element $x$ belongs to the neutrosophic population $M$ in the following way, $x(t, i, f) \in M$, which means: $x$ is $t$ % in the population $M$, $f$ %$x$ is not in the population $M$, while $i$ % the appurtenance of $x$ to $M$ is indeterminate (unknown, unclear, neutral: neither in the population nor outside).

**Example.** Let's consider the population of a country $C_1$. Most people in this country have only the citizenship of the country, therefore they belong 100% to $C_1$. But there are people that have double citizenships, of countries $C_1$ and $C_2$. Those people belong 50% to $C_1$, and 50% to $C_2$. While citizens with triple citizenships of countries $C_1$, $C_2$, and $C_3$ belong only 33.33% to each country. Of course, considering various criteria these percentages may differ. Also, there are countries with autonomous zones, whose citizens in these zones may not entirely consider themselves as belonging to those countries.

But there is another category of people that have been stripped from their $C_1$ citizenship for political reasons and they have other citizenship, while still living (temporarily) in $C_1$.They are called *paria*, and they do not belong to $C_1$ (not having citizenship), but still belong to $C_1$ (because they still living in $C_1$). They form the indeterminate part of neutrosophic population of country $C_1$.

A **simple random neutrosophic sample of size *n*** from a classical or neutrosophic population is a sample of *n* individuals such that at least one of them has some indeterminacy.

**Example.** One considers a random sample of 1,000 homes, in a city of over one million inhabitants, in order to investigate how many houses have at least a laptop. One finds out that 600 houses have at least one laptop, 300 houses don't have any laptop, while 100 houses have each of them a single laptop, but not working.

Some of these 100 house owners tried to have their laptop fixed, others said their laptops' hard drives have crashed and it is little chance to fix them. Therefore indeterminacy. We have a simple random neutrosophic sample of size 100.

Similarly as in classical statistics, in a **stratified random neutrosophic sampling** the pollster groups the (classical or neutrosophic) population by a strata according to a classification; afterwards the pollster takes a random sample (of appropriate size according to a criterion) from each group. If there is some indeterminacy, we deal with neutrosophic sampling.

**Example.** One considers two strata: men and women in the city of Gallup, New Mexico. But, since women represent 51% of the population and men 49%, one takes a random sample of 51 women and a random sample of 49 men.

But later learn that „one" man and two "women" are actually transgender. Therefore 3 individuals are indeterminate. Whence one has stratified random neutrosophic sampling.

If the (classical or neutrosophic) population is divided into subgroups, such that each subgroup is representative of the population, and then one collects from these subgroups a random sample and there is some indeterminacy, then one has a **neutrosophic cluster sampling**.

**Example.** Suppose 5 professors conduct PhD dissertations in neutrosophic statistics. Each professor has a number of graduate students, but some students are undecided whether to pursue their dissertations in classical or neutrosophic statistics. The professors represent the clusters. One randomly selects 2 professors to interview their students about research in neutrosophic statistics. But, because some students are undecided (indeterminate) with respect to their research topic, we have a neutrosophic cluster sampling.

A **convenience sample** is likely to be inaccurate since the pollster selects a sample of individuals that are readily available, who might answer randomly to the questions in order to finish faster. The less the individuals are interested in the survey results, the more likely inaccurate are the survey results. While a **voluntary-response sample** is more likely to be biased, since the sample individuals may volunteer in purpose to influence the survey results.

Besides these two categories of sample individuals there is another one of *malicious people* that might oppositely answer to the questions in order to produce false results.

That's why data of some sample individuals has to be removed, but often we don't know which ones. Therefore, we have indeterminacy related to the sample size: how many sample people were from the above three categories, and how to depict their data in order to remove them from the survey results? Again, *neutrosophic statistics.*

# Neutrosophic Numerical Measures

Example with Neutrosophic Numbers $a + bI$, where $a$, $b$ are real numbers, and $I$ is indeterminacy, such that $I^2 = I$ and $0 \cdot I = 0$.

Let's have the neutrosophic numbers:
$-2 - 4I$, $-1 + 0 \cdot I$, $3 + 5I$, $6 + 7I$.
Compute their *mean*:

$$\frac{(-2 - 4I) + (-1 + 0 \cdot I) + (3 + 5I) + (6 + 7I)}{4} =$$

$$= \frac{-2 - 1 + 3 + 6}{4} + \frac{-4 + 0 + 5 + 7}{4} \cdot I = 1.5 + 2I.$$

Compute their *median*:

$$\frac{(-1 + 0 \cdot I) + (3 + 5I)}{2} = \frac{-1 + 3}{2} + \frac{0 + 5}{2}I = 1 + 2.5I.$$

Compute the *deviation* of each neutrosophic number with respect to the mean:

$$(-2 - 4I) - (1.5 + 2I) = -3.5 - 6I,$$
$$(-1 + 0 \cdot I) - (1.5 + 2I) = -2.5 - 2I,$$
$$(3 + 5I) - (1.5 + 2I) = 1.5 + 3I,$$
$$(6 + 7I) - (1.5 + 2I) = 4.5 + 5I.$$

Square the deviations:

$$(-3.5 - 6I)^2 = (-3.5)^2 + 2(-3.5)(-6)I + (-6)^2 I^2$$
$$= 12.25 + 42I + 36I^2 = 12.25 + 42I + 36I$$
$$= 12.25 + 78I$$
$$(-2.5 - 2I)^2 = 6.25 + 14I$$
$$(1.5 + 3I)^2 = 2.25 + 18I$$
$$(4.5 + 5I)^2 = 20.25 + 70I.$$

We are following the formula:

$$(a + bI)^2 = a^2 + 2abI + b^2 I^2$$
$$= a^2 + 2abI + b^2 I$$

or

$$(a + bI)^2 = a^2 + (2ab + b^2)I.$$

Compute the **standard deviation**:

$$s =$$

$$= \sqrt{\frac{(12.25 + 78I) + (6.25 + 14I) + (2.25 + 18I) + (20.25 + 70I)}{4}}$$

$$= \sqrt{10.25 + 45I}.$$

To compute the square root of a neutrosophic number we denote the result as $x + yI$ and determine $x$ and $y$:

$$\sqrt{10.25 + 45I} = x + yI.$$

Raise both sides to the second power:

$$10.25 + 45I = x^2 + (2xy + y^2)I.$$

Therefore:

$$\begin{cases} 10.25 = x^2 \\ 45 = 2xy + y^2 \end{cases}$$

Since standard deviation is positive, we take

$$x = +\sqrt{10.25} \simeq 3.20$$

and replace it into the second equation:

$$45 = 2(3.20)y + y^2$$

and solve for positive y:

$$y^2 + 6.4y - 45 = 0$$

whence

$$y = \frac{-6.4 + \sqrt{6.4^2 - 4(1)(-45)}}{2(1)} \simeq 0.64.$$

Therefore, the neutrosophic standard deviation of the previous four neutrosophic numbers is

$$3.20 + 0.64I.$$

We observe that 3.20 is the classical standard deviation of the determinate parts of the previous neutrosophic numbers: $-2, -1, 3, 6$; but 0.64 is not the classical standard deviation of the indeterminate parts of the previous neutrosophic numbers: $-4, 0, 5, 7$.

The classical standard deviation of the numbers -4, 0, 5, 7, whose mean is 2, is:

$$\sqrt{\frac{(-4-2)^2 + (0-2)^2 + (5-2)^2 + (7-2)^2}{4}} \simeq 4.30.$$

Indeterminacy has propagated when squaring the deviations.

# Classical Neutrosophic Numbers

A classical Neutrosophic Number has the standard form:

$$a + bI,$$

where $a$, $b$ are real or complex coefficients, and $I$ = indeterminacy, such $0 \cdot I = 0$ and $I^2 = I$.

It results that $I^n = I$ for all positive integer $n$.

If the coefficients $a$ and $b$ are real, then $a + bI$ is called **Neutrosophic Real Number**.

Examples: $2 + 3I, -5 + \frac{7}{3}I$, etc.

But if the coefficients $a$ and $b$ be are complex, then $a + bI$ is called **Neutrosophic Complex Number**.

Examples: $(5 + 2i) + (2 - 8i)I, I + i + 9I - iI$, etc. where $i = \sqrt{-1}$.

A neutrosophic complex number can be better written as:

$a + bi + cI + diI$, where $a$, $b$, $c$, and $d$ are reals.

Of course, any real number can be considered, by language abuse, a neutrosophic number.

For example:

$$5 = 5 + 0 \cdot I,$$

or

$5 = 5 + 0 \cdot i + 0 \cdot I + 0 \cdot i \cdot I.$

We call it a degenerated neutrosophic number.

A true neutrosophic number contains the indeterminacy $I$ with a non-zero coefficient.

**Division of classical neutrosophic real numbers.**

$$(a_1 + b_1I) \div (a_2 + b_2I) = ?$$

We denote the result by:

$$\frac{a_1 + b_1 I}{a_2 + b_2 I} = x + yI,$$

then multiply and identify the coefficients:

$$a_1 + b_1 I \equiv (x + yI)(a_2 + b_2 I)$$
$$\equiv xa_2 + xb_2 I + ya_2 I + yb_2 I^2$$
$$\equiv (a_2 x) + (b_2 x + a_2 y + b_2 y)I.$$

Whence we form an algebraic system of equations, by identifying the coefficients:

$$a_2 x = a_1$$
$$b_2 x + a_2 y + b_2 y = b_1$$

or

$$a_2 x = a_1$$
$$b_2 x + (a_2 + b_2)y = b_1.$$

One obtains unique solution only when the determinant of second order $\begin{vmatrix} a_2 & 0 \\ b_2 & a_2 + b_2 \end{vmatrix} \neq 0$

or $a_2(a_2 + b_2) \neq 0$. Hence $a_2 \neq 0$ and $a_2 \neq -b_2$ are that conditions for the division of neutrosophic real numbers

$$\frac{a_1 + b_1 I}{a_2 + b_2 I}$$

to exist.

Then

$$x = \frac{a_1}{a_2}$$

and

$$y = \frac{a_2 b_1 - a_1 b_2}{a_2(a_2 + b_2)}$$

or

$$\frac{a_1 + b_1 I}{a_2 + b_2 I} = \frac{a_1}{a_2} + \frac{a_2 b_1 - a_1 b_2}{a_2(a_2 + b_2)} \cdot I.$$

As consequences, we have:

1. $\frac{a+bI}{ak+bkI} = \frac{a+bI}{k(a+bI)} = \frac{1}{k}$, for k a non-zero real number, and for $a \neq 0$ and $a \neq -b$.

2. $\frac{I}{a+bI} = \frac{a}{a(a+b)} \cdot I = \frac{1}{a+b} \cdot I$, for $a \neq 0$ and $a \neq -b$.

3. Divisions by I, -I, and in general by $kI$, for $k$ a real, are undefined.

$\frac{a+bI}{kI}$ = undefined, for any real $k$, and any reals $a$ and $b$.

In particular:

$$\frac{I}{I} = \text{undefined};$$

$$\frac{7I}{I} = \text{undefined};$$

$$\frac{10I}{5I} = \text{undefined};$$

$$\frac{a+bI}{I} = \text{undefined};$$

$$\frac{a+bI}{-I} = \text{undefined}.$$

4. $\frac{a+bI}{c} = \frac{a}{c} + \frac{b}{c} \cdot I$, for $c \neq 0$;

5. $\frac{c}{a+bI} = \frac{c}{a} - \frac{bc}{a(a+b)} \cdot I$, for $a \neq 0$ and $a \neq -b$.

6. $\frac{a+0 \cdot I}{b+0 \cdot I} = \frac{a}{b}$, for $b \neq 0$ (the classical division of reals).

7. $\frac{a+bI}{1} = \frac{a}{1} + \frac{b}{1} \cdot I = a + bI.$

8.  $\frac{0}{a+b\cdot I} = \frac{0}{a} + \frac{a\cdot 0 - 0\cdot b}{a(a+b)}\cdot I = 0 + 0\cdot I = 0,$   for   $a \neq 0$ and $a \neq -b$.

9.  $\frac{kI}{a+bI} = \frac{k}{a+b}\cdot I$, for any real $k$, and $a \neq 0$ and $a \neq -b$.

Let's fo a **concrete example** by calculation.

What is $(2 + 3I) \div (1 + I) = ?$

Denote:

$$\frac{2 + 3I}{1 + I} = x + yI.$$

One has:

$$(1 + I)(x + yI) = x + yI + xI + yI^2 \equiv 2 + 3I$$
$$x + (x + 2y)I \equiv 2 + 3I.$$

Whence $\begin{cases} x = 2 \\ x + 2y = 3 \end{cases}$

or $x = 2, y = 0.5$.

There

$$\frac{2 + 3I}{1 + I} = 2 + 0.5I.$$

Let's check:

$$\frac{2 + 3I}{2 + 0.5I} = x + yI.$$

Then

$$(2 + 0.5I)(x + yI) \equiv 2 + 3I,$$
$$2x + (2y + 0{,}5x + 0.5y) \equiv 2 + 3I.$$

Whence

$$\begin{cases} 2x = 2 \\ 0.5x + 2.5y = 3' \end{cases}$$

hence

$$x = 1, \qquad y = 1,$$

or

$$\frac{2 + 3I}{2 + 0.5I} = 1 + 1 \cdot I = 1 + I.$$

Perfect.

*Another example.*

$$\frac{2 + 3I}{8 + 12I} = x + yI.$$

Whence

$$\begin{cases} 8x = 2 \\ 12x + 12y + 8y = 3 \end{cases}$$

and we get

$$x = \frac{2}{8} = \frac{1}{4},$$

and

$$12 \left(\frac{1}{4}\right) + 20y = 3, \quad \text{or } y = 0.$$

Therefore

$$\frac{2 + 3I}{8 + 12I} = \frac{1}{4} + 0 \cdot I = \frac{1}{4},$$

which is a neutrosophic simplification since:

$$\frac{2 + 3I}{8 + 12I} = \frac{1 \cdot (2 + 3I)}{4 \cdot (2 + 3I)} = \frac{1}{4}.$$

Now *an example which is undefined*:

$$\frac{2 + 3I}{1 - I} = ?$$

$$\frac{2 + 3I}{1 - I} = x + yI$$

$$(1 - I)(x + yI) \equiv 2 + 3I$$

$$x + yI - xI - yI^2 \equiv 2 + 3I$$

or

$$x + (y - x - y)I \equiv 2 + 3I$$

or

$$x - xI \equiv 2 + 3I,$$

therefore

$$x = 2 \quad \text{and} \quad -x = 3,$$

which is impossible.

Therefore

$$\frac{2 + 3I}{1 - I} = \text{undefined.}$$

And an example where it results infinitely solutions:

$$\frac{I}{I} = ?$$

Denote

$$\frac{I}{I} = x + yI,$$

so

$$I(x + yI) \equiv I,$$

or

$$xI + yI^2 \equiv I,$$

or

$$(x + y)I \equiv 1 \cdot I,$$

whence $x + y = 1$, where $x$ and $y$ are unknown reals.

We get infinitely many solutions:

$x \in \mathcal{R}$ and $y = 1 - x$,

where $\mathcal{R}$ is the set of real numbers. Among solutions there are:

1, $I$, 2-$I$, etc.

But since the division's result should be unique, we say that

$$\frac{I}{I} = \text{undefined.}$$

## Root index $n \geq 2$ of a neutrosophic real number.

First let's compute the square root:
$\sqrt{a + bI}$, where $a$, $b$ are reals.
Let's denote:

$$\sqrt{a + bI} = x + yI,$$

where x and y are real unknowns, and raise both sides to the second power. One gets:

$$a + bI \equiv (x + yI)^2 = x^2 + 2xyI + y^2I^2 = x^2 + 2xyI + y^2I$$
$$= x^2 + (2xy + y^2)I.$$

Whence $\begin{cases} x^2 = a \\ 2xy + y^2 = b. \end{cases}$

Hence $\begin{cases} x = \pm\sqrt{a} \\ y^2 \pm 2\sqrt{a} \cdot y - b = 0 \end{cases}$

and we solve the second equation for y:

$$y = \frac{\mp 2\sqrt{a} \pm \sqrt{4a + 4b}}{2(1)} = \frac{\mp 2\sqrt{a} \pm 2\sqrt{a + b}}{2}$$
$$= \mp\sqrt{a} \pm \sqrt{a + b},$$

and the four solution are:

$$(x, y) = (\sqrt{a}, -\sqrt{a} + \sqrt{a + b}), (\sqrt{a}, -\sqrt{a} - \sqrt{a + b}),$$
$$(-\sqrt{a}, \sqrt{a} + \sqrt{a + b}), \text{or} (-\sqrt{a}, \sqrt{a} - \sqrt{a + b}).$$

Thus:

$$\sqrt{a + bI} = \sqrt{a} + (-\sqrt{a} + \sqrt{a + b})I,$$

or

$$\sqrt{a} - (\sqrt{a} + \sqrt{a + b})I,$$

or

$$-\sqrt{a} + \left(\sqrt{a} + \sqrt{a+b}\right)I,$$

or

$$-\sqrt{a} + \left(\sqrt{a} - \sqrt{a+b}\right)I.$$

Let's consider an example done through all calculations:

$$\sqrt{9 + 7I} = ?$$

Let's denote:

$$\sqrt{9 + 7I} = x + yI.$$

Then:

$$9 + 7I = x^2 + 2xyI + y^2I^2 = x^2 + (2xy + y^2)I.$$

Whence

$$\begin{cases} x^2 = 9, \text{or } x = \pm 3 \\ 2xy + y^2 = 7 \end{cases}.$$

Let's find $y$:

$$x = 3 \qquad x = -3$$
$$6y + y^2 = 7 \qquad -6 + y^2 = 7$$
$$y^2 + 6y - 7 = 0 \quad y^2 - 6y - 7 = 0$$
$$(y+7)(y-1) = 0 \quad (y-7)(y+1) = 0$$
$$y = -7/y = 1 \quad y = 7/y = -1$$
$$(3,-7),(3,1) \quad (-3,7),(-3,-1).$$

Therefore, $\sqrt{9 + 7I} = \pm 3 \pm I$ (four solutions).

As a **particular case** we can compute $\sqrt{I}$.

Let's consider $\sqrt{I} = x + yI$, then

$$0 + 1 \cdot I = x^2 + (2xy + y^2) \cdot I$$

and we need to find $x$ and $y$.

Whence $x^2 = 0$, or $x = 0$,

and $2xy + y^2 = 1$, or $y^2 = 1$, or $y = \pm 1$.

Hence $\sqrt{I} = \pm I$.

Similarly for $\sqrt[n]{I}$.

Let's consider $\sqrt[n]{I} = x + yI$,

or $0 + 1 \cdot I = x^n + \left( \sum_{k=0}^{n-1} C_n^k y^{n-k} x^k \right) \cdot I$,

where $x^n = 0$, or $x = 0$,

and

$$\sum_{k=0}^{n-1} C_n^k y^{n-k} x^k = 1,$$

or $y^n = 1$, whence $y = \sqrt[n]{1}$ and we get $n$ solutions: a real solution $y = 1$ and $n - 1$ complex solutions in the case we are interested in neutrosophic complex solutions as roots index $n$ of 1.

In the same way, we can compute root index $n \geq 2$ of any neutrosophic number:

$$\sqrt[n]{a - bI} = x + yI$$

or

$$a + bI = (x + yI)^n$$
$$= x^n + \left( y^2 + \sum_{k=0}^{n-1} C_n^k y^{n-k} x^k \right) \cdot I =$$
$$= x^n + \left( \sum_{k=0}^{n-1} C_n^k y^{n-k} x^k \right) \cdot I,$$

where $C_n^k$ means combination of $n$ elements taken by groups of $k$ elements.

Whence $x = \sqrt[n]{a}$ if $n$ is odd, or $x = \pm\sqrt[n]{a}$ if $n$ is even,

and

$$\left( \sum_{k=0}^{n-1} C_n^k y^{n-k} a^{\frac{k}{n}} \right) = b,$$

and solve it for $y$.

When the x and $y$ solutions are real, we get neutrosophic real solutions, and when x and y solutions are complex, we get neutrosophic complex solutions.

Let $a + bi + cI + diI$ be **a neutrosophic complex number**, where $a, b, c, d$ are reals. Let's compute square root of it:

$$\left(\sqrt{a + bi + cI + diI}\right)^2 = (x + yi + zI + wiI)^2$$

$$
\begin{aligned}
a + bi &+ cI + diI \\
&= x^2 - y^2 + z^2I^2 + w^2i^2I^2 + 2xyi + 2xzI \\
&\quad + 2xwiI + 2yziI + 2ywi^2I + 2zwiI^2 \\
&= x^2 - y^2 + z^2I - w^2I + 2xyi + 2xzI \\
&\quad + 2xwiI + 2yziI - 2ywI + 2zwiI \\
&= (x^2 - y^2) + 2xyi \\
&\quad + (z^2 - w^2 + 2xz - 2yw)I \\
&\quad + (2xw + 2yz + 2zw)iI.
\end{aligned}
$$

Then we get a non-linear algebraic system in four variables ($x, y, z, w$) and four equations:

$$
\begin{cases}
x^2 - y^2 = a \\
2xy = b \\
z^2 - w^2 + 2xz - 2yw = c \\
2xw + 2yz + 2zw = d.
\end{cases}
$$

In a more general way, we can compute **root index** $n$ of a neutrosophic complex number:

$$(a + bi + cI + diI)^{\frac{1}{n}} = x + yi + zI + wiI,$$

where $x, y, z, w$ are variables in the set of real numbers.

Raising to the power n in both sides, one gets:

$$a + bi + cI + diI = (x + yi + zI + wiI)^n$$
$$= f_1(x, y) + f_2(x, y)i + f_3(x, y, w, z)I$$
$$+ f_4(x, y, w, z)iI,$$

where $f_1, f_2, f_3, f_4$ are real functions.

Whence we get a non-linear algebraic system in four variables ($x$, $y$, $w$, $z$) and four equations:

$$\begin{cases} f_1(x, y) = a \\ f_2(x, y) = b \\ f_3(x, y, w, z) = c \\ f_4(x, y, w, z) = d, \end{cases}$$

that we need to solve.

Similarly, one can compute **square root of a complex number**.

Let $a + bi$, where $i = \sqrt{-1}$, and a, b are reals, be a complex number.

$\sqrt{a + bi} = x + yi$ such that $(x + yi)^2 \equiv a + bi$,

where $x$ and $y$ are real numbers;

or $x^2 + 2xyi + y^2i^2 \equiv a + bi$,

or $(x^2 - y^2) + (2xy)i \equiv a + bi$,

whence $\begin{cases} x^2 - y^2 = a \\ 2xy = b. \end{cases}$

From the first equation $x = \pm\sqrt{y^2 + a}$ is replaced into the second equation:

$\pm 2y\sqrt{y^2 + a} = b$. (RE)

Raising both sides to the second power one gets:

$$4y^2(y^2 + a) = b^2,$$

or

$$4y^4 + 4ay^2 - b^2 = 0.$$

Let $z = y^2$. Then $4z^2 + 4az - b^2 = 0$, then

$$z = \frac{-4a \pm \sqrt{16a^2 - 4\cdot 4(-b^2)}}{2(4)} = \frac{-4a \pm \sqrt{16a^2 + 16b^2}}{8}$$
$$= \frac{-4a \pm 4\sqrt{a^2 + b^2}}{8} = \frac{-a \pm \sqrt{a^2 + b^2}}{2}.$$

Then

$$y = \pm\sqrt{\frac{-a \pm \sqrt{a^2 + b^2}}{2}},$$

and

$$x = \frac{b}{2y} = \frac{b}{\pm 2\sqrt{\frac{-a\pm\sqrt{a^2+b^2}}{2}\cdot\frac{2}{2}}} = \frac{\pm b}{\sqrt{2a \pm 2\sqrt{a^2 + b^2}}},$$

for $y \neq 0$.

Since (RE) is a radical equation, we need to check each solution of unknown $y$ to make sure the solution is not extraneous.

Because $\sqrt{a^2 + b^2} \geq \pm a$, the expression $\frac{-a+\sqrt{a^2+b^2}}{2} \geq$ 0, therefore there are at least two real values for $y$,

$$y = \pm\sqrt{\frac{-a + \sqrt{a^2 + b^2}}{2}},$$

while $-a - \sqrt{a^2 + b^2} \leq 0$ and one has equality only when $b = 0$, resulting in $y = 0$.

As a *particular case*, $\sqrt{i} = \frac{\sqrt{2}}{2} + \frac{\sqrt{2}}{2}\cdot i$, or $-\frac{\sqrt{2}}{2} - \frac{\sqrt{2}}{2}i$,

since we write:

$i = 0 + 1\cdot i$, whence $a = 0$, $b = 1$,

and we replace both of them into the $x$ and $y$ of previous formulas.

We can check the results:

$$\left(\frac{\sqrt{2}}{2} + \frac{\sqrt{2}}{2} \cdot i\right)^2 = \frac{2}{4} + 2 \cdot \frac{2}{4}i + \frac{2}{4}i^2 = \frac{1}{2} + i - \frac{1}{2} = i,$$

and similarly $\left(-\frac{\sqrt{2}}{2} - \frac{\sqrt{2}}{2}i\right)^2 = i$.

Let's have another example, doing all calculations:

$$\sqrt{3 - 4i} = ?$$

Denote $\quad \sqrt{3 - 4i} = x + yi$.

Then

$$3 - 4i \equiv (x + yi)^2 = (x^2 - y^2) + (2xy)i.$$

Whence

$$\begin{cases} x^2 - y^2 = 3 \\ 2xy = -4. \end{cases}$$

Solve this system.

From the second equation, $y = \frac{-2}{x}$, and replace $y$ into the first:

$$x^2 - \left(\frac{-2}{x}\right)^2 = 3,$$

or

$$x^2 - \frac{4}{x^2} - 3 = 0,$$

or

$$x^4 - 3x^2 - 4 = 0,$$

or

$$(x^2 - 4)(x^2 + 1) = 0,$$

whence

$$x^2 - 4 = 0,$$

or

$$x = \pm 2.$$

Then

$$y = \frac{-2}{x} = \frac{-2}{\pm 2} = \mp 1.$$

Solutions:

$$\sqrt{3 - 4i} = \pm(2 - i).$$

Checking the result:

$$[\pm(2 - 1)]^2 = 4 - 4i + i^2 = 3 - 4i.$$

Remarkably, we'll get the same solutions if we take the complex values of $x$ and $y$, because:

$$x^2 + 1 = 0 \text{ gives } x = \pm\sqrt{-1} = \pm i,$$

and replacing them into the substitution $y = \frac{-2}{x}$, we get:

$$y = \frac{-2}{\pm i} = \frac{-2}{\pm i} \cdot \frac{i}{i} = \frac{-2i}{\pm i^2} = \frac{-2i}{\mp i} = \pm 2i.$$

Then

$$\sqrt{3 - 4i} = x + yi = \pm i \pm 2i \cdot i = \pm i \pm 2(-1) = \mp 2 \pm i$$
$$= \pm(2 - i).$$

One generalizes this procedure and one computes **root index $n$ of any complex number**:

$$\sqrt[n]{a + bi} = ?$$

Similarly denote:

$$\sqrt[n]{a + bi} = x + yi,$$

then

$$a + bi \equiv (x + yi)^n = (yi + x)^n$$
$$= \sum_{k=0}^{\left[\frac{n}{2}\right]} C_n^{2k} y^{2k} i^{2k} x^{n-2k}$$
$$+ \sum_{k=0}^{\left[\frac{n-1}{2}\right]} C_n^{2k+1} y^{2k+1} i^{2k+1} x^{n-2k-1},$$

and one obtains a non-linear algebraic system of degree $n$, in two variables $x$ and $y$, and two equations:

$$\begin{cases} \displaystyle\sum_{k=0}^{\left[\frac{n}{2}\right]} C_n^{2k} y^{2k} (-1)^k x^{n-2k} = a \\ \displaystyle\sum_{k=0}^{\left[\frac{n-1}{2}\right]} C_n^{2k+1} y^{2k+1} (-1)^k x^{n-2k-1} = b \end{cases}$$

that one solves with a computer program.

As a *particular case*, let's compute the cubic root of a complex number:

$$\sqrt[3]{a + bi} = ?$$

Then:

$$\sqrt[3]{a + bi} = x + yi,$$

or

$$a + bi \equiv (x + yi)^3 = x^3 + 3x^2 yi + 3xy^2 i^2 + y^3 i^3$$
$$= (x^3 - 3xy^2) + (3x^2 y - y^3)^i,$$

whence

$$\begin{cases} x^3 - 3xy^2 = a \\ 3x^2 y - y^3 = b \end{cases}$$

and solve for $x$ and $y$.

From the first equation:

$$y = \pm\sqrt{\frac{x^3 - a}{3x}}.$$

Replace this substitution into the second equation:

$$\pm 3x^2 \sqrt{\frac{x^3 - a}{3x}} \mp \left(\sqrt{\frac{x^3 - a}{3a}}\right)^3 - b = 0$$

and solve this independent equation for x with a calculator, and then find y from the above substitution.

For example:

$$\sqrt[3]{i} = -i.$$

## Neutrosophic Real or Complex Polynomial.

A polynomial whose coefficients (at least one of them containing *I*) are neutrosophic numbers is called **Neutrosophic Polynomials**.

Similarly we may have **Neutrosophic Real Polynomials** if its coefficients are neutrosophic real numbers, and **Neutrosophic Complex Polynomials** if its coefficients are neutrosophic complex numbers.

Examples:

$$P(x) = x^2 + (2 - I)x - 5 + 3I$$

is a neutrosophic real polynomial, while

$$Q(x) = 3x^3 + (1 + 6i)x^2 + 5Ix - 4iI$$

is a neutrosophic complex polynomial.

From these polynomials we proceed to solving **Neutrosophic Real** or **Complex Polynomial Equations**.

Let's consider the following neutrosophic real polynomial equation:

$$6x^2 + (10 - I)x + 3I = 0,$$

and solve it just using the quadratic fromula:

$$x = \frac{-(10 - I) \pm \sqrt{(10 - I)^2 - 4(6)(3I)}}{2(6)}$$

$$= \frac{-10 + I \pm \sqrt{100 - 20I + I^2 - 72I}}{12}$$

$$= \frac{-10 + I \pm \sqrt{100 - 20I + I - 72I}}{12}$$

$$= \frac{-10 + I \pm \sqrt{100 - 91I}}{12}$$

Now, we need to compute $\sqrt{100 - 91I}$.

Let's denote: $\sqrt{100 - 91I} = \alpha + \beta I$,

where α, β are reals.

Raise both sides to the second power:

$$100 - 91I = \alpha^2 + 2\alpha\beta I + \beta^2 I^2 = \alpha^2 + 2\alpha\beta I + \beta^2 I$$
$$= \alpha^2 + (2\alpha\beta + \beta^2)I,$$

whence

$$\begin{cases} \alpha^2 = 100 \\ 2\alpha\beta + \beta^2 = -91. \end{cases}$$

Hence $\alpha = \pm\sqrt{100} = \pm 10$.

1. If $\alpha = 10$, then $2(10)\beta + \beta^2 = -91$, or $\beta^2 + 20\beta + 91 = 0$. Using the quadratic formula, one gets:

$$\beta = \frac{-20 \pm \sqrt{20^2 - 4(1)91}}{2} = \frac{-20 \pm \sqrt{400 - 364}}{2}$$

$$= \frac{-20 \pm 6}{2} = \langle \begin{matrix} \frac{-20 + 6}{2} = -7; \\ \frac{-20 - 6}{2} = -13. \end{matrix}$$

2.  If $\alpha = -10$, then $\beta^2 - 20\beta + 91 = 0$,

$$\beta = \frac{20 \pm \sqrt{(-20)^2 - 4(1)91}}{2} = \frac{20 \pm \sqrt{400 - 364}}{2}$$

$$= \frac{20 \pm 6}{2} = \langle \begin{matrix} \frac{20 + 6}{2} = 13; \\ \frac{20 - 6}{2} = 7. \end{matrix}$$

The four solutions are:

$(\alpha, \beta) = (10, -7), (10, -13), (-10, 13), (-10, 7).$

We go back now and find $x$:

$$x = \frac{-10 + I \pm \sqrt{100 - 9 + I}}{12}.$$

Therefore, we previously found out that $\sqrt{100 - 91I} = 10 - 7I$, or $-10 + 7I$, or $10 - 13I$, or $-10 + 13I$.

Since one has $\pm$ in front of the radical, $10 - 7i$ and $-10 + 7I$ get the same values for $x$. Similarly, $10 - 13I$ and $-10 + 13I$.

$$x_{1,2} = \frac{-10 + I \pm (10 - 7I)}{12}$$

$$= \langle \begin{matrix} \frac{-10 + I + 10 - 7I}{12} = \frac{-6I}{12} = -\frac{1}{2}I; \\ \frac{-10 + I - 10 + 7I}{12} = \frac{-20 + 8I}{12} = -\frac{5}{3} + \frac{2}{3}I; \end{matrix}$$

$$x_{1,2} = \frac{-10 + I \pm (10 - 13I)}{12}$$

$$= \langle \begin{array}{l} \dfrac{-10 + I + 10 - 13I}{12} = \dfrac{-12I}{12} = -I; \\[2mm] \dfrac{-10 + I - 10 + 13I}{12} = \dfrac{-20 + 14I}{12} = -\dfrac{10}{6} + \dfrac{7}{6}I. \end{array}$$

We got four neutrosophic solutions

$\left\{ -\dfrac{1}{2}I, -\dfrac{5}{3} + \dfrac{2}{3}I, -I, -\dfrac{10}{6} + \dfrac{7}{6}I \right\}$ for

a neutrosophic real polynomial of degree 2.

First neutrosophic factoring:

$P(x) = 6x^2 + (10 - I)x + 3I = 6\left[x - \left(-\dfrac{1}{2}I\right)\right] \cdot \left[x - \left(-\dfrac{5}{3} + \dfrac{2}{3}I\right)\right] = 6\left(x + \dfrac{1}{2}I\right)\left(x + \dfrac{10}{6} - \dfrac{7}{5}I\right).$

Second neutrosophic factoring:

$$P(x) = 6x^2 + (10 - I)x + 3I$$
$$= 6[x - (-I)] \cdot \left[x - \left(-\frac{10}{6} + \frac{7}{6}I\right)\right]$$
$$= 6\left(x + \frac{1}{2}I\right)\left(x + \frac{10}{6} - \frac{7}{5}I\right).$$

Differently from the classical polynomial with real or complex coefficients, **the neutrosophic polynomials do not have a unique factoring**!

If we check each solution, we get:

$$P(x_1) = P(x_2) = P(x_3) = P(x_4) = 0.$$

Let's compute:

$$P(x_4) = P\left(\frac{-10}{6} + \frac{7}{6}I\right)$$

$$= 6 \cdot \left(\frac{-10}{6} + \frac{7}{6}I\right)^2 + (10 - I)\left(\frac{-10}{6} + \frac{7}{6}I\right) + 3I$$

$$= 6\left(\frac{100}{36} - \frac{140}{36}I + \frac{49}{36}I^2\right) + \left(\frac{-100}{6}\right) + \frac{70}{6}I$$

$$+ \frac{10I}{6} - \frac{7}{6}I^2 + 3I$$

$$= \frac{100}{6} - \frac{140 \cdot I}{6} + \frac{49 \cdot I}{6} - \frac{100}{6} + \frac{70 \cdot I}{6} + \frac{10 \cdot I}{6}$$

$$- \frac{7 \cdot I}{6} + \frac{18I}{6}$$

$$= \frac{-140I + 49I + 70I + 10I - 7I + 18I}{6} = \frac{0 \cdot I}{6}$$

$$= \frac{0}{6} = 0.$$

Another procedure of factoring a neutrosophic real polynomial equation is the following.

Let's have

$$P(x) = (A + B \cdot I)x^2 + (C + D \cdot I)x + (E + F \cdot I) = 0.$$

Suppose $x_1 = a_1 + b_1 I$ and $x_2 = a_2 + b_2 I$ are two neutrosophic real solutions of $P(x) = 0$.

Then:

$$P(x) = (A + B \cdot I)[x - (a_1 + b_1 I)] \cdot [x - (a_2 + b_2 I)]$$
$$\equiv (A + B \cdot I)x^2 + (C + D \cdot I)x$$
$$+ (E + F \cdot I).$$

We multiply on the second right hand side, and then we identify the neutrosophic coefficients, and solve for $a_1, b_1, a_2$ and $b_2$.

### Research Problems.

1. In general, how many neutrosophic solutions has a neutrosophic real polynomial equation of degree $n \geq 1$?

So far, we know that such equation of degree 1 has none (in the case the neutrosophic division is undefined) or one solution (in the case the neutrosophic division is well defined).

2. How many different factorings, with factors of first degree, are possible for a neutrosophic real polynomial of degree *n*? We got two different factorings for aparticular polynomial of degree 2.

3. – 4. Similar problems for neutrosophic complex polynomial equations and neutrosophic complex polynomials of degree $n \geq 1$.

# Neutrosophic Random Numbers

**Neutrosophic Random Numbers** can also be generated using, instead of only crisp numbers, a pool of sets. For example, let's suppose one has 100 balls and on each ball is written an interval $[a, b]$ where $a, b \in \{1, 2, 3, …, 100\}$ and $a \leq b$.

When $a = b$ we get a crisp number $[a, a] = a$, while for $a < b$ we get a set *[a, b]*.

Then randomly one extracts a ball, one registers its interval, then one returns it back to the pool. And so on. Instead of a random sequence of crisp numbers, we get a random sequence of intervals.

# Example with Neutrosophic Data

Let's have the following four observations:

$$6, [2, 5], 30, [18, 24].$$

The second and fourth observations are unclear, i.e. [2,5] means a number in this interval, but we don't know which one; similarly for [18, 24]. Therefore we have two indeterminacies.

In order to uniformize let's rewrite all observations as intervals:

$$[6, 6], [2, 5], [30, 30], [18, 24].$$

Each observations can be a subset, not necessarily a crisp number a (closed, open, half closed – half open) interval.

Compute the *median*:

$$\frac{[2, 5] + [30, 30]}{2} = \frac{[2 + 30, 5 + 30]}{2} = \frac{[32, 35]}{2} = \left[\frac{32}{2}, \frac{35}{2}\right]$$
$$= [16, 17.5].$$

Therefore the medium is a number between 16 and 17.5.

One computes their mean:

$$\frac{[6, 6] + [2, 5] + [30, 30] + [18, 24]}{4}$$
$$= \frac{[6 + 2 + 30 + 18, 6 + 5 + 30 + 24]}{4}$$
$$= \frac{[56, 65]}{4} = \left[\frac{56}{4}, \frac{65}{4}\right] = [14, 16.25].$$

Therefore the average is a number between 14 and 16.25.

Compute the deviations and square them:

a. $[6, 6] - [14, 16.2] = [6 - 16.2, 6 - 14] = [10.2, -8]$;

$$[10.2, -8]^2 = [-10.2, -8] \cdot [-10.2, -8]$$
$$= [(-8)(-8), (-10.2) \cdot (-10.2)]$$
$$= [64, 104.04].$$

b. $[2, 5] - [14, 16.25] = [2 - 16.25, 5 - 14] = [-14, 25, -9]$;

$$[-14.25, -9]^2 = [(-9)^2, (-14.25)^2] = [81, 203.0625];$$

c. $[30, 30] - [14, 16.2] = [30 - 16.2, 30 - 14] = [13.8, 16]$;

$$[13.8, 16]^2 = [13.8^2, 16^2] = [190.44, 256];$$

d. $[18, 24] - [14, 16.2] = [18 - 16.2, 24 - 14] = [1.8, 10]$;

$$[1.2, 10]^2 = [1.8^2, 10^2] = [3.24, 100].$$

Compute the **standard deviation**:

$$\sqrt{\frac{[64, 104.04] + [81, 203.0625] + [190.44, 256] + [3.24, 100]}{4}}$$
$$=$$
$$= \sqrt{\left[\frac{64 + 81 + 190.44 + 3.24}{4}, \frac{104.04 + 203.0625 + 256 + 100}{4}\right]} =$$
$$= \sqrt{[84.67, 165.775625} =$$
$$\left[\sqrt{84.67}, \sqrt{165.775625}\right] \simeq [9.20163, 12.8754].$$

# Indeterminacy related to the sample size

Suppose one has the following five observations:

$$17, 12, 5, 8, 9,$$

but one of them is certainly wrong, yet we don't know which one.

What to do to approximate the calculations?

Let's first increasable reorder the observations:

$$5, 8, 9, 12, 17,$$

and then study all possibilities.

| Sample Number | Wrong Observations | Correct Observations | Median | Mean | Deviations | Squared Deviations | Standard Deviation |
|---|---|---|---|---|---|---|---|
| 1 | | 8 | | | −3.5 | 12.25 | |
| | | 9 | | | −2.5 | 6.25 | |
| | | 12 | | | 0.5 | 0.25 | |
| | | 17 | | | 5.5 | 30.25 | |
| | 5 | | 10.5 | 11.5 | | | 3.5 |
| 2 | | 5 | | | −5.75 | 33.0625 | |
| | | 9 | | | −1.75 | 3.0625 | |
| | | 12 | | | 1.25 | 1.5625 | |
| | | 17 | | | 6.25 | 39.0625 | |
| | 8 | | 10.5 | 10.75 | | | 4.38035 |
| 3 | | 5 | | | −5.5 | 30.25 | |
| | | 8 | | | −2.5 | 6.25 | |
| | | 12 | | | 1.5 | 2.25 | |
| | | 17 | | | 6.5 | 42.25 | |
| | 9 | | 10.0 | 10.5 | | | 4.5 |
| 4 | | 5 | | | −4.75 | 22.5625 | |
| | | 8 | | | −1.75 | 3.0625 | |
| | | 9 | | | −0.75 | 0.5625 | |
| | | 17 | | | 7.25 | 52.5625 | |
| | 12 | | 8.5 | 9.75 | | | 4.43706 |
| 5 | | 5 | | | −3.5 | 12.25 | |
| | | 8 | | | −0.5 | 0.25 | |
| | | 9 | | | 0.5 | 0.25 | |
| | | 12 | | | 3.5 | 12.25 | |
| | 17 | | 8.5 | 8.5 | | | 2.5 |

Now we combine the five results.

### a. Interval style:

$$\begin{cases} \text{the median belongs to the interval } [8.5, 10.5]; \\ \text{the mean belongs to the interval } [8.5, 11.5]; \\ \text{the standard deviation belongs to the interval } [2.5, 4.43706]. \end{cases}$$

### b. Average Style:

$$the\ median = \frac{10.5 + 10.5 + 10.0 + 8.5 + 8.5}{5} = 9.6;$$

$$the\ mean = \frac{11.5 + 10.75 + 10.5 + 9.75 + 8.5}{5} = 10.2;$$

and $standard\ deviation$

$$= \frac{3.5 + 4.38035 + 4.5 + 4.43706 + 2.5}{5}$$

$$\simeq 3.86348.$$

### c. Weight Average Style:

One assigns a weight to each sample. The sample weight may represent the chance that the respective sample could be the right sample, after discarding the wrong observations.

In general, the weights $w_1, w_2, \ldots, w_n \in [0, 1]$ such that

$$w_1 + w_2 + \cdots + w_n = 1.$$

In the case when the sample weights are determined from criteria different from each other and therefore the sum of weights is not 1, and the observations are $a_1 + a_2 + \cdots + a_n$, the weight average is:

$$\frac{w_1 a_1 + w_2 a_2 + \cdots + w_n a_n}{w_1 + w_2 + \cdots + w_n}.$$

In our example, if $w_1 = 0.4, w_2 = 0.1, w_3 = 0.3, w_4 = 0.2, w_5 = 0.7$, then:

$the weighted average median$

$$= \frac{0.4(10.5) + 0.1(10.5) + 0.3(10.0) + 0.2(8.5) + 0.7(8.5)}{0.4 + 0.1 + 0.3 + 0.2 + 0.7}$$

$\simeq 9.35294;$

$the weighted average mean$

$$= \frac{0.4(11.5) + 0.1(10.75) + 0.3(10.5) + 0.2(9.75) + 0.7(8.5)}{0.4 + 0.1 + 0.3 + 0.2 + 0.7}$$

$\simeq 9.83824$

and $the weighted average deviation$

$$= \frac{0.4(3.5) + 0.1(4.38035) + 0.3(4.5) + 0.2(4.43706) + 0.7(2.5)}{0.4 + 0.1 + 0.3 + 0.2 + 0.7}$$

$\simeq 3.42673.$

According to the sample weights, it's a larger chance that the right sample is the fifth one. Therefore, the combined statistical metrics of all samples would be inclined to approach the fifth sample's statistical metrics.

This example can be generalized for $n$ observations, such that $k$ observations among them are wrong, where $n \geq 2$ and $1 \leq k \leq n - 1$.

With a computer program, one studies each of the $C_n^{n-k}$ samples resulted after discarding $k$ wrong observations, where $C_n^{n-k}$ means combinations of $n$ elements taken in groups of $n$-$k$ elements. Each sample has the size $n$-$k$. For each sample one calculates its median, mean, deviations, standard deviations, and of course other statistical metrics required by the neutrosophic problem to solve.

Then we combine $C_n^{n-k}$ results using either interval style, the average style, the weighted average style, or other procedures that the reader may design depending on the problem.

# Neutrosophic Binomial Distribution

The classical Binomial Distribution is extended neutrosophically. That means that there is some indeterminacy related to the probabilistic experiment.

Suppose each trial can result in an outcome labeled success (*S*), or its mutually exclusive outcome labeled failure (*F*), or some indeterminacy (*I*).

For example: tossing a coin on an irregular surface which has cracks, the coin can fall inside a crack on its edge, and thus one gets neither head, nor tale, but indeterminacy.

We conduct a fixed number of small experiments (that we call *trials*). The outcomes of the trials are independent. For each trial, the chance of getting *S* is the same; similarly for the chance of getting *F*, or of getting *I*.

The **neutrosophic binomial random variable** *x* is then defined as the number of successes when we perform the experiment $n \geq 1$ times.

The **neutrosophic probability distribution of *x*** is also called **neutrosophic binomial probability distribution**.

For *n* trials it is important the way one defines the indeterminacy.

First, it is clear that getting indeterminacy in each trial means indeterminacy for the whole set of $n$ trials. Secondly, getting indeterminacy in no trial means no indeterminacy for the whole set of $n$ trials.

But what about getting indeterminacy in some trials, and determinacy (i.e. success or failure) in other trials?

This partially indeterminate and partially determinate set of $n$ trials depends on the problem one needs to solve and on the expert's point of view.

One can define an **indeterminacy threshold**:

$th$ = number of trials whose outcome is indeterminate,

$$\text{where } th \in \{0, 1, 2, \dots, n\}.$$

The cases with a $threshold > th$ will belong to the indeterminate part, while for a $threshold \leq th$ they will belong to the determinate part.

Let *P(S)* = the chance a particular trial results in a success,

and  *P(F)* = the chance a particular trial results in a failure, for both *S* and *F* different from indeterminacy.

Let *P(I)* = the chance a particular trial results in an indeterminacy.

For  $x \in \{0, 1, 2, \dots, n\}$, $NP$ (exactly x successes among $n$ trials) = $(T_x, I_x, F_x)$, with

$$T_x = \frac{n!}{x!\,(n-x)!} \cdot P(S)^x \cdot \sum_{k=0}^{th} C_{n-x}^k\, P(I)^k P(F)^{n-x-k}$$

$$= \frac{n!}{x!\,(n-x)!} \cdot P(S)^x$$

$$\cdot \sum_{k=0}^{th} \frac{(n-x)!}{(n-x-k)!} P(I)^k P(F)^{n-x-k}$$

$$= \frac{n!}{x!} \cdot P(S)^x \cdot \sum_{k=0}^{th} \frac{P(I)^k P(F)^{n-x-k}}{k!\,(n-x-k)!}.$$

Similarly:

$$F_x = \sum_{\substack{y=0 \\ y \neq x}}^{n} T_y = \sum_{\substack{y=0 \\ y \neq x}}^{n} \frac{n!}{y!} \cdot P(S)^y \cdot \left[\sum_{k=0}^{th} \frac{P(I)^k \cdot P(F)^{n-y-k}}{k!\,(n-y-k)!}\right], \text{and}$$

$$I_x = \sum_{z=th+1}^{n} \frac{n!}{z!\,(n-z)!} \cdot P(I)^z$$

$$\cdot \left[\sum_{k=0}^{n-z} C_{n-z}^k P(S)^k \cdot P(F)^{n-z-k}\right]$$

$$= \sum_{z=th+1}^{n} \frac{n!}{z!\,(n-z)!} \cdot P(I)^z$$

$$\cdot \left[\sum_{k=0}^{n-z} \frac{(n-z)!}{k!\,(n-z-k)!} P(S)^k \cdot P(F)^{n-z-k}\right]$$

$$= \sum_{z=th+1}^{n} \frac{n!}{z!} \cdot P(I)^z$$

$$\cdot \left[\sum_{k=0}^{n-z} \frac{P(S)^k \cdot P(F)^{n-z-k}}{k!\,(n-z-k)!}\right],$$

where $C_u^v$ means combinations of $u$ elements taken by groups of $v$ elements:

$$C_u^v = \frac{u!}{v!\,(u-v)!}$$

and $u!$ is $u$ factorial, $u! = 1 \cdot 2 \cdot 3 \cdot \ldots \cdot u$.

Also:

$T_x$ = chance of $x$ successes, and $n - x$ failures and indeterminacies but such that the number of indeterminacies is less than or equal to indeterminacy threshold;

$F_x$ = chance of $y$ successes, with $y \neq x$ and $n - y$ failures and indeterminacies but such that the number of indeterminacies is less than or equal to the indeterminacy threshold;

and $I_x$ = chance of $z$ indeterminacies, where $z$ is strictly greater than the indeterminacy threshold.

$$T_x + I_x + F_x = (P(S) + P(I) + P(F))^n.$$

In most applications,

$$P(S) + P(I) + P(F) = 1,$$

and this case is called **complete probability**.

But for **incomplete probability** (where there is missing information):

$$0 \leq P(S) + P(I) + P(F) < 1.$$

While in the **paraconsistent probability** (which has contradictory information):

$$1 < P(S) + P(I) + P(F) \leq 3.$$

### An Example.

Among the watches sold by a store 80% had a digital display and 10% an analog display. There is a number of watches sold for which the storeowner has no evidence about their type of display, and he asks his manager assistant about them. Not knowing the manager's previous estimations, the assistant estimates the unknown type of watches to be 20%.

Let's consider a neutrosophic random variable

$x$ = the number of watches among the next 5 buyers that have an analog display.

Therefore:

$$P(F) = P(digital display) = 0.8,$$
$$P(S) = P(analog display) = 0.1,$$
$$P(I) = P(indeterminacy) = 0.2.$$

We got a paraconsistent neutrosophic probability since the information comes from the different sources that estimate independently. We have contradiction between the estimations of the manager and his assistant, because

$$0.8 + 0.1 + 0.2 = 1.1 > 1.$$

We have a neutrosophic binomial distribution.

Let's say the indeterminacy threshold is 2.

We define the random variable $X$ as follows:

$x$ = number of watches that have an analog display among the next 5 watches to be bought;

$$T_x = \frac{5!}{x!}(0.1)^x \cdot \sum_{k=0}^{2} \frac{(0.2)^k (0.8)^{5-x-k}}{k!\,(5-x-k)!},$$

where $x = 0, 1, 2, 3, 4, 5.$

The chance that exactly 2 watches are analog, i.e. $x = 2$ is:

$$T_2 = \frac{5!}{2!}(0.1)^2 \cdot \left[\frac{(0.2)^0 \cdot (0.8)^3}{0! \, 3!} + \frac{(0.2)^1 (0.8)^2}{1! \, 2!}\right.$$
$$\left. + \frac{(0.2)^2 (0.8)^1}{2! \, 1!}\right] = 0.0992.$$

$$I_2 = \sum_{z=2+1}^{5} \frac{5!}{z!} \cdot (0.2)^z \cdot \left[\sum_{k=0}^{5-z} \frac{(0.1)^k (0.8)^{5-z-k}}{k! \, (5-z-k)!}\right]$$
$$= \frac{5!}{3!}(0.2)^3$$
$$\cdot \left[\sum_{k=0}^{2} \frac{(0.1)^k (0.8)^{2-k}}{k! \, (2-k)!}\right] (for \, z = 3)$$
$$+ \frac{5!}{4!}(0.2)^4 \cdot \left[\sum_{k=0}^{1} \frac{(0.1)^k (0.8)^{1-k}}{k! \, (1-k)!}\right] (for \, z = 4) +$$
$$\frac{5!}{5!}(0.2)^5 \cdot \left[\sum_{k=0}^{0} \frac{(0.1)^k (0.8)^{1-k}}{k! \, (-k)!}\right] (for \, z = 5)$$
$$= 20 \cdot (0.2)^3 \cdot \left[\frac{(0.1)^0 (0.8)^2}{0! \, 2!} + \frac{(0.1)^1 (0.8)^1}{1! \, 1!} + \frac{(0.1)^2 (0.8)^0}{2! \, 0!}\right]$$
$$+ 5 \cdot (0.2)^4 \cdot \left[\frac{(0.1)^0 (0.8)^1}{0! \, 1!} + \frac{(0.1)^1 (0.8)^0}{1! \, 0!}\right]$$
$$+ 1 \cdot (0.2)^5 \cdot \left[\frac{(0.1)^0 (0.8)^0}{0! \, 0!}\right] = 0.07232.$$

$F_2$ can easier be computed (instead of using its combinatorial formula) as :
$$F_2 = (P(S) + P(I) + P(F))^5 - T_2 - I_2$$
$$= (0.1 + 0.2 + 0.8)^5 - 0.0992 - 0.07232$$
$$= 1.43899.$$

If we normalize the vector
$$(T_2, I_2, F_2) = (0.0992, 0.07232, 1.43899)$$

by dividing each vector component by their total sum

0.0992 + 0.07232 + 1.43899 = 1.61051,

we get $(T_2, I_2, F_2) = (0.061595, 0.044905, 0.893500)$.

For incomplete and paraconsistent probabilities it doesn't matter if we normalize at the beginning or at the end, we'll get the same result.

*

**Remark**.

Since a third component (the chance of indeterminacy) was added to the binomial distribution, the neutrosophic binomial distribution actually resembles a summation of classical trinomial distribution:

$$(p_1 + i + p_2)^n$$

where $p_1$ and $p_2$ are the probabilities that the two mutually exclusive events ( $E_1$ and $E_2$ ) occur respectively, while «$i$» is the chance of getting indeterminacy.

Let's denote by $A(\alpha, \beta, \gamma)$ the probability of obtaining $a$ events $E_1$, β indeterminate events, and γ events $E_2$, where of course $0 \le \alpha, \beta, \gamma \le n$, and $\alpha + \beta + \gamma = n$, as results of $n$ independent trials.

Of course, as in classical trinomial distribution, one has

$$A(\alpha, \beta, \gamma) = \frac{n!}{\alpha! \, \beta! \, \gamma!} \cdot p_1^\alpha i^\beta p_2^\gamma$$

with n = $\alpha + \beta + \gamma$.

We need to define what indeterminacy means within $n$ trials. Let $th$ be the indeterminacy threshold. For $th + 1$ or more indeterminacies, we consider them as indeterminacy, otherwise we have determinacy.

Then for $x \in \{0, 1, 2, \dots, n\}$,

$NP$(exactly $x$ events $E$, among $n$ trials) $= (T_x, I_x, F_x)$, where :

$$T_x = \sum_{0 \leq \beta \leq th} A(x, \beta, n - x - \beta)$$

$$I_x = \sum_{\substack{th+1 \leq \beta \leq n \\ 0 \leq \alpha \leq n-th}} A(\alpha, \beta, n - \alpha - \beta)$$

$$F_x = \sum_{\substack{0 \leq \alpha \leq n, \ \alpha \neq x \\ 0 \leq \beta \leq th}} A(\alpha, \beta, n - \alpha - \beta)$$

# Neutrosophic Multinomial Distribution

The previous neutrosophic binomial distribution is generalized for the case when at each trial there are $r \, (\geq 2)$ possible outcomes and some indeterminacy.

Suppose all possible outcomes are
$$E_1, E_2, \ldots, E_r$$
with corresponding chances to occur
$$P_1, P_2, \ldots, P_r$$
and some indeterminacy $I$ with corresponding chance to occur $i$.

Then we have the multinomial expansion:
$$(P_1 + P_2 + \cdots + P_r + i)^n$$
for $n$ trials.

Let's denote similarly by $A(\alpha_1, \alpha_2, \ldots, \alpha_r, \beta)$ the probability of obtaining: exactly $\alpha_1$ events $E_1$, $\alpha_2$ events $E_2$, … , $\alpha_r$ events $E_r$, and $\beta$ indeterminate events,

where $0 \leq \alpha_1, \, \alpha_2, \ldots, \alpha_r, \beta \leq n$
and $\alpha_1 + \alpha_2 + \cdots + \alpha_r + \beta = n,$

as results of $n$ independent trials, then
$A(\alpha_1, \, \alpha_2, \ldots, \alpha_r, \beta)$
$$= \frac{n!}{\alpha_1! \; \alpha_2! \; \ldots \; \alpha_r! \; \beta} \cdot P_1^{\alpha_1} \cdot P_2^{\alpha_2} \cdot \ldots \cdot P_r^{\alpha_r} \cdot i^{\beta}.$$

Consider the same $th$ as indeterminacy treshold.

Let the random variable $X_j$ denotes the number of times events $E_j$ occurs, for any $j \in \{1, 2, \dots, r\}$, in $n$ independent trials.

So we have a **multivariate distribution**.

Then the neutrosophic probability of obtaining exactly $x_1$ events $E_1$, $x_2$ events $E_2$, ..., $x_r$ events $E_r$, in n trials is

$\left( T_{x_1, x_2, \dots, x_r}, \ I_{x_1, x_2, \dots, x_r}, F_{x_1, x_2, \dots, x_r} \right)$, where

$$T_{x_1, x_2, \dots, x_r} = \sum_{0 \le \beta \le th} A(x_1, \ x_{2, } \dots, x_r, \beta)$$

$$I_{x_1, x_2, \dots, x_r} = \sum_{\substack{th+1 \le \beta \le n \\ 0 \le \alpha_j \le n-th, \ \text{for } j \in \{1,2,\dots,r\}}} A(\alpha_1, \alpha_{2, } \dots, \alpha_r, \beta)$$

$$F_{x_1, x_2, \dots, x_r} =$$
$$\sum_{\substack{0 \le \beta \le th \\ (\alpha_1, \alpha_2, \dots, \alpha_r) \in \{1,2,\dots,n\}^r \setminus (x_1, x_2, \dots, x_r)}} A(\alpha_1, \alpha_{2, } \dots, \alpha_r, \beta).$$

# Neutrosophic Scatter Plot

A **Neutrosophic Scatter Plot** is a picture of points (*x, y*), such that at least a point is not well defined.

For example the point (3, 5) is well defined, while the points ([2, 4], 7) or (-6, [0, 1]) or ({−2, −4}, 3) or ([1, 2], [5, 7]) are imprecise.

As an example, let's consider a sample of size *n* = 4 yielding the accompanying data:

| Neutrosophic Observation | | | | |
|---|---|---|---|---|
| | **1** | **2** | **3** | **4** |
| ***x*** | 2 | 4 | $[5, 6]$ | 3 |
| ***y*** | $[1, 2]$ | 3 | $[3, 4]$ | 5 |

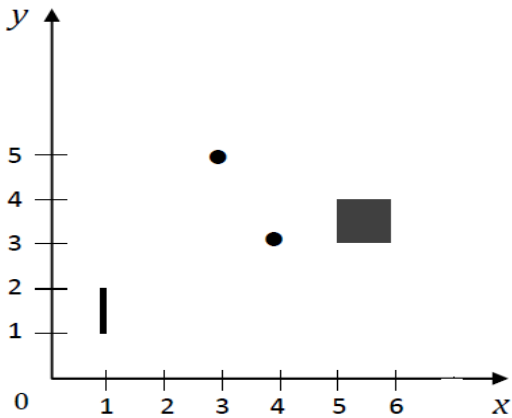

2D NEUTROSOPHIC SCATTER PLOT

The **bivariate neutrosophic scatter plot** has, besides points as in classical scatter plot, also segment of lines or parts of segments of lines, or surfaces, or parts of surfaces (geometrical objects of dimensions 1 or 2).

In general, an ***n*-variate neutrosophic scatter** plot formed by $n - 1$ independent variables and one dependent variable, is composed of geometrical objects of dimensions 0, 1, 2, …, or $n$.

A **neutrosophic dependent or response variable** is a dependent variable that has some indeterminacy.

Similarly, a **neutrosophic independent** or **predictor variable** is a variable that has some indeterminacy.

A **neutrosophic function**
$$f_N(x_1, x_2, …, x_n) = 0$$
is a function depending on variables $x_1, x_2, …, x_n$ such that the function has at least one indeterminate coefficient, or at least one of its independent variables $x_1, x_2, …, x_n$ has some indeterminate value or is unknown.

Indeterminate coefficient or indeterminate value can be a subset with two or more elements.

The graph of a neutrosophic function in general has a higher dimension than the graph of a corresponding classical function (whose indeterminacies have been removed).

For examples, the classical function $f(x, y) = 0$ represents a curve in the 2D-space, while the neutrosophic function $f_N(x, y) = 0$ can be a surface.

The classical function $f(x, y, z) = 0$ represents a surface in 3D-space, while the neutrosophic function $f_N(x, y, z) = 0$ can represent a bigger surface or a solid.

And in general while a classical function

$$f(x_1, x_2, ..., x_n) = 0$$

is a geometrical object of dimension $d$ in the $n$-dimensional space, a neutrosophic function

$$f_N(x_1, x_2, ..., x_n) = 0$$

is a bigger (as volume) geometrical object of dimension d, or a geometrical object of dimension > d.

The study of a neutrosophic function becomes more difficult when, for example, a function's coefficient or a value of one of its independent variables is completely unknown.

More classical statistical formulas can be neutrosophically extended by replacing the operations on crisp numbers with operations on sets, that we present below.

Let's $S_1$ and $S_2$ be two sets of numbers.
Then:

$S_1 + S_2 = \{x_1 + x_2 | x_1 \in S_1 \text{ and } x_2 \in S_2\}$(set addition)

$S_1 - S_2 = \{x_1 - x_2 | x_1 \in S_1 \text{ and } x_2 \in S_2\}$(set substraction)

$S_1 \cdot S_2 = \{x_1 \cdot x_2 | x_1 \in S_1 \text{ and } x_2 \in S_2\}$(set multiplication)

$a \cdot S_1 = S_1 \cdot a = \{a \cdot x_1 | x_1 \in S_1\}$(scalar multiplication)

$$a + S_1 = S_1 + a = \{a + x_1 | x_1 \in S_1\}(\text{scalar set addition})$$
$$a - S_1 = \{a - x_1 | x_1 \in S_1\}(\text{scalar set substraction})$$
$$S_1 - a = \{x_1 - a | x_1 \in S_1\}(\text{scalar set substraction})$$
$$\frac{S_1}{S_2} = \left\{\frac{x_1}{x_2} \Big| x_1 \in S_1, x_2 \in S_2, x_2 \neq 0 \right\}(\text{set division})$$
$$S_1^n = \{x_1^n | x_1 \in S_1\}(\text{set power})$$
$$\frac{S_1}{a} = \left\{\frac{x_1}{a} \Big| x_1 \in S_1, a \neq 0 \right\}(\text{set scalar division})$$
$$\frac{a}{S_1} = \left\{\frac{a}{x_1} \Big| x_1 \in S_1, x_1 \neq 0 \right\}(\text{set scalar division})$$
$$\sqrt[n]{S_1} = \{\sqrt[n]{x_1} | x_1 \in S_1\} \quad (\text{root index}n\text{of a set})$$

As generalizations wehave:
$$\sum_{i=1}^{m} S_i = \{\textstyle\sum_{i=1}^{m} x_i \, | x_i \in S_i \text{for all} i = 1, 2, \dots, m\}.$$
Similarly :
$$\prod_{i=1}^{m} S_i = \{\textstyle\prod_{i=1}^{m} x_i \, | x_i \in S_i \text{for all} i = 1, 2, \dots, m\}.$$

# Neutrosophic Regression

Neutrosophic Regression is the analysis of the association between one or more independent variables and a dependent variable that are expressed by neutrosophic values. This association is usually formulated as a neutrosophic equation or formula, which enables prediction of future values of the dependent variable.

The graph of this association is, instead of a curve in classical statistics, for example:

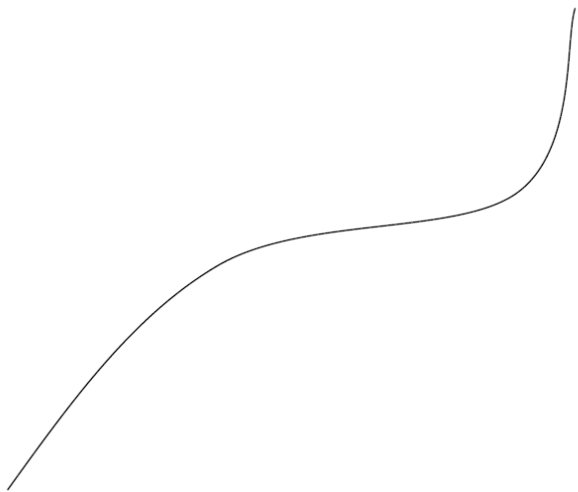

a neutrosophic curve (we can call it a „thick curve", or „strip curve"), like:

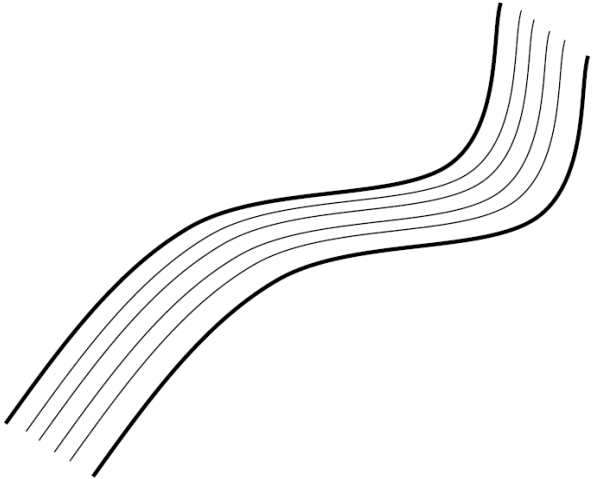

since in neutrosophic theory one deals with indeterminacy and approximations.

As in classical statistics, the neutrosophic regression may be *linear* (if the association between independent, and dependent variables is linear), or *non-linear* (if the association is non-linear). Among the neutrosophic non-linear regressions of second degree one mentions the parabolic, elliptic, and hyperbolic regressions.

# Neutrosophic Least-Squares Lines

The **Neutrosophic Least-Squares Lines** that approximates the neutrosophic bivariate data
$$(x_1, y_1), (x_2, y_2), \dots , (x_n, y_n)$$
has the same formula as in classical statistics
$$\hat{y} = a + by$$
where the slope
$$b = \frac{\sum xy - [(\sum x)(\sum y)/n]}{\sum x^2 - [(\sum x)^2/n]}$$
and the $y$ – intercept
$$a = \bar{y} - b\bar{x}$$
with $\bar{x}$ the neutrosophic average of $x$,
and $\bar{y}$ the neutrosophic average of $y$.

One uses the circumflex accent ^ above $y$ in order to emphasize that $\hat{y}$ is a prediction of $y$.

The only distinction from classical least-square line is that in neutrosophic theory we work with sets instead of numbers.

Therefore, into the data, some $x$'s or $y$'s are imprecise, expressed by sets. The consequence is that « a » or « b » could result in being sets instead of numbers.

Let's see an example.

| Neutrosophic Observation Number | $x$ | $y$ | $x^2$ | $xy$ | $y^2$ | Neutrosophic Predicted Value $\hat{y}i$ | Neutrosophic Residual $yi, \hat{y}i$ |
|---|---|---|---|---|---|---|---|
| 1 | 2 | [1, 3] | 4 | [2, 6] | [1, 9] | (-21.3587, 18.7955) | (-17.7985, 24.3587) |
| 2 | [4, 5] | 6 | [16, 25] | [24, 30] | 36 | (-20.5014, 38.5603) | (-32.5603, 26.5014) |
| 3 | 1 | 2 | 1 | 2 | 4 | (-21.7871, 12.2073) | (-10.2073, 23.7871) |
| 4 | (6, 7) | (10, 13) | (36, 49) | (60, 91) | (100, 169) | (-19.6443, 51.7367) | (-41.7367, 32.6443) |
| 5 | 8 | {14, 15} | 64 | {112, 120} | {196, 225} | (-18.7871, 58.325) | (-44.325, 33.7871) |
| 6 | 3 | 5 | 9 | 15 | 25 | (-20.93, 25.3838) | (-20.3838, 25.93) |
| Sum | (24, 26) | (38, 44) | (130, 152) | (215, 264) | (362, 468) | | |
| | ↑ | ↑ | ↑ | ↑ | ↑ | | |
| | $\sum x$ | $\sum y$ | $\sum x^2$ | $\sum xy$ | $\sum y^2$ | | |

TABLE OF A NEUTROSOPHIC SAMPLE

An example of calculation with sets:

$$\sum y = [1, 3] + 6 + 2 + (10, 13) + 5 + \{14, 15\}$$
$$= (1 + 6 + 2 + 10 + 5, 3 + 6 + 2 + 13 + 5)$$
$$+ \{14, 15\} = (24, 29) + \{14, 15\}$$
$$= \{(24, 29) + 14, (24, 29) + 15\}$$
$$= \{(38, 43), (39, 44)\} = (38, 44).$$

Whence:

$$b = \frac{(215, 264) - [(24, 26) \cdot \frac{(38,44)}{6}]}{(130, 152) - [\frac{(24,26)^2}{6}]}$$

$$= \frac{(215, 264) - [\frac{912,1144}{6}]}{(130, 152) - [\frac{576,676}{6}]}$$

$$\simeq \frac{(215, 264) - (152, 191)}{(130, 152) - (96, 113)} = \frac{(24, 112)}{(17, 56)}$$

$$= \left(\frac{24}{56}, \frac{112}{17}\right) \simeq (0.42857, 6.58824).$$

Since

$$\bar{x} = \frac{(24, 26)}{6} \simeq (4, 4.33333)$$

and

$$\bar{y} = \frac{(38, 44)}{6} = \left(\frac{38}{6}, \frac{44}{6}\right) \simeq (6.33333, 7.33333)$$

we get

$a = (6.33333, 7.33333) - (0.42857, 6.58824) \cdot$
$(4, 4.33333) = (6.33333, 7.33333) - (1.71428, 28.549) =$
$(-22.2157, 5.61905)$.

Thus, the neutrosophic least-squares line is :

$\hat{y} = (-22.2157, 5.61905) + (0.42857, 6.58824)x$.

Let's graph this «line», which actually is a geometrical surface between two lines.

If $x = 0, \hat{y} = (-22.2157, \ 5.61905)$.

If $x = 1, \hat{y} = (-22.2157 + 0.42857, \ 5.61905 + 6.58824) = (-21.7871, \ 12.2073)$.

We plot these neutrosophic points, which are actually segments of line.

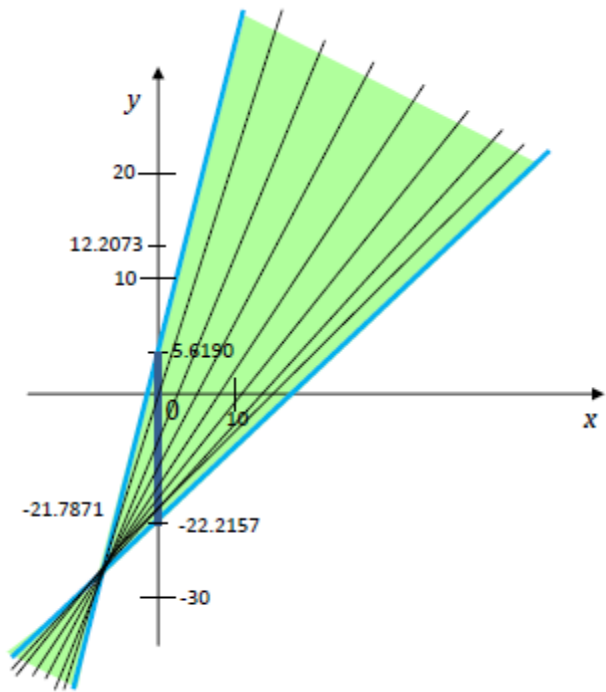

*Neutrosophic Predicted Values* are computed as
$$\hat{y}i = (-22.2157, 5.61905) + (0.42857, 6.58824)x_i,$$
$$\text{for } i = 1, 2, \dots, 6.$$

Hence:
$$\widehat{y_1} = (-22.2157, 5.61905) + (0.42857, 6.58824) \cdot 2$$
$$= (-22.2157 + 0.4285 \cdot 2, 5.61905$$
$$+ 6.58824 \cdot 2) = (-21.3587, 18.7955).$$

$$\widehat{y_2} = (-22.2157, 5.61905) + (0.42857, 6.58824) \cdot [4, 5]$$
$$= (-22.2157 + 0.42857 \cdot 4,$$
$$5.61905 + 6.58824 \cdot 5)$$
$$= (-20.5014, 38.5603).$$
$$\widehat{y_3} = (-22.2157, 5.61905) + (0.42857, 6.58824) \cdot 1$$
$$= (-22.2157 + 0.42857, 5.61905$$
$$+ 6.58824 \cdot 1) = (-21.7871, 12.2073).$$
$$\widehat{y_4} = (-22.2157, 5.61905) + (0.42857, 6.58824) \cdot (6, 7)$$
$$= (-22.2157 + 0.42857 \cdot 6, 5.61905$$
$$+ 6.58824 \cdot 7) = (-19.6443, 51.7367).$$
$$\widehat{y_5} = (-22.2157, 5.61905) + (0.42857, 6.58824) \cdot (8)$$
$$= (-22.2157 + 0.42857 \cdot 8, 5.61905$$
$$+ 6.58824 \cdot 8) = (-18.7871, 58.325).$$
$$\widehat{y_6} = (-22.2157, 5.61905) + (0.42857, 6.58824) \cdot 3$$
$$= (-22.2157 + 0.42857 \cdot 3, 5.61905$$
$$+ 6.58824 \cdot 3) = (-20.93, 25.3838).$$

The **Neutrosophic Residuals** are computed in the same way as in classical statistics:

$$y_1 - \widehat{y_1}, y_2 - \widehat{y_2}, \dots, y_n - \widehat{y_n}$$

where $y_i$ are the real values of variable y,
and $\widehat{y_i}$ are respectively their predicted values.
The neutrosophic residuals are:

$$y_1 - \widehat{y_1} = [1, 3] — [(22.2157, 5.61905)$$
$$+ (0.42857, 6.58824) \cdot 2 ]$$
$$= [1, 3] - (21.3587, 18.7955)$$
$$= (1 - 18.7955, 3 - (-21.3587))$$
$$= (-17.7955, 24.3587).$$
$$y_2 - \widehat{y_2} = 6 - [(-22.2157, 5.61905) + (0.42857, 6.58824)$$
$$\cdot [4, 5]) = 6 - (-20.5014, 38.5603)$$
$$= (-32.5603, 26.5014)$$

$y_3 - \widehat{y_3} = 2 — [(-22.2157, 5.61905) + (0.42857, 6.58824)$
$\cdot 1] = 2 — (-21.7871, 12.2073)$
$= (-10.2073, 23.7871).$

$y_4 - \widehat{y_4} = (10, 13) - [(-22.2157, 5.61905) + (0.42857, 6.58824) \cdot (6, 7)] = (10, 13) - (-19.6443, 51.7367) = (-41.7367, 32.6443).$

$y_5 - \widehat{y_5} = \{14, 15\} - [(-22.2157, 5.61905) + (0.42857, 6.58824) \cdot 8] = \{14, 15\} - (18.7871, 58.325) = (-44.325, 33.7871).$

$y_6 - \widehat{y_6} = 5 - [(-22.2157, 5.61905) + (0.42857, 6.58824)$
$\cdot 3] = 5 - (-20.93, 25.3838)$
$= (-20.3838, 25.93).$

It is remarkable to observe that each real value of belongs to or it is included in the predicted value interval:

$$y_1 = [1, 3] \subset (-21.3587, 18.3955);$$
$$y_2 = 6 \in (-20.5014, 38.5603);$$
$$y_3 = 2 \in (-21.7871, 12.2073);$$
$$y_4 = (10, 13) \subset (-19.6643, 51.7367);$$
$$y_5 = \{14, 15\} \subset (-18.7871, 58.325);$$
$$y_6 = 5 \in (-20.93, 25.3838).$$

## Deneutrosofications.

a. Another idea of solving this problem would be to transform the neutrosophic data in classical data, either taking the midpoint of each set, or the average of a discrete set of the form {…}. Or taking small neighborhoods centered in the midpoints of each set. Or taking the minimum values of the sets and thus constructing multiple classical data. Then one computes the least-squares line for each data. Afterwards one makes the average of the results, or one considers the min/max interval of the results.

b. Or one transforms the neutrosophic least-squares line into a classical least-square line by replacing the set representations of the coefficients «a» and «b» by their corresponding midpoint, or (depending on the application) by other interior points of the two sets. In our previous example,

$\hat{y} = (-22.2157, 5.61905) + (0.42857, 6.58824) \cdot x$

becomes

$$\hat{y} = -8 + 3.5x,$$

where $-8$ is close to the mipoint of $(-22.2157, 5.61905)$,

and $3.5$ is close to the midpoint of $(0.42857, 6.58824)$.

c. One could take the midpoints of the neutrosophic predicted values neutrosophic residuals, or initial neutrosphic data; or smaller neighborhoods centered in the midpoints; or *min* values and *max* values separately and obtaining multiple classical data and calculating the needed statistical characteristic for each of them, then averaging the results.

Let's compute the midpoints of neutrosophic predicted values and neutrosophic residuals:

| Neutrosophic Predicted Value Midpoint | Neutrosophic Residual Midpoint |
|---|---|
| -1.2816 | 3.2801 |
| 9.0295 | -3.0295 |
| -4.7899 | 6.7899 |
| 16.0467 | -4.5462 |
| 19.7690 | -5.2690 |
| 2.2269 | 2.7731 |

# Neutrosophic Coefficient of Determination

We compute the **Neutrosophic Residual Sum of Squares**, denoted by *NSSResid*, given by:

$$NSSResid = \sum (y - \hat{y})^2 = \sum y^2 - a \sum y - b \sum xy$$

and the **Neutrosophic Total Sum of Squares**, denoted by

$$NSSTo = \sum (y - \bar{y})^2 = \sum y^2 - \frac{(\sum y)^2}{n}.$$

The **Neutrosophic Coefficient of Determination**, denoted by $r_N^2$, is :

$$r_N^2 = 1 - \frac{NSSResid}{NSSTo},$$

and represents the proportion of variation in y, when considering a linear relationship between variables $x$ and $y$.

$$\begin{aligned}
NSSResid &= 3.2801^2 + (3.0295)^2 + 6.7899^2 \\
&\quad + (-4.5462)^2 + (-5.2690)^2 + (2.7731)^2 \\
&= 122.16;
\end{aligned}$$

$$\begin{aligned}
NSSTo &= \sum y^2 - \frac{(\sum y)^2}{n} = (362, 468) - \frac{(38, 44)^2}{6} \\
&= (362, 468) - \left( \frac{38^2}{6}, \frac{44^2}{6} \right) \\
&= (362, 468) - (40.1111, 53.7778) \\
&= (362 - 53.7778, 468 - 40.1111) \\
&= (308.222, 427.889).
\end{aligned}$$

Whence

$$r_N^2 = 1 - \frac{122 \cdot 16}{(308.222, 327.889)} = 1 - \left(\frac{122 \cdot 16}{327.889}, \frac{122.16}{308.222}\right)$$
$$= 1 - (0.3726, 0.3963)$$
$$= (1 - 0.3963, 1 - 0.3726)$$
$$= (0.6037, 0.6274).$$

So between 60.37% and 62.74% of the sample variation is explained by the neutrosophic approximate linear relationship between $x$ and $y$.

The Neutrosophic Correlation Coefficient or the product moment neutrosophic coefficient $r_N$ (extension of Pearson's correlation coefficient from crisp data to neutrosophic data), has the same formula as in classical statistics, but we work with sets instead of numbers:

$$r_N = \frac{n \sum xy - \sum x \sum y}{\sqrt{[n \sum x^2 - (\sum x)^2][n \sum y^2 - (\sum y)^2]}}$$

or

$$r_N = \frac{S_{xy}}{S_x S_y},$$

where $S_{xy}$ is the neutrosophic covariance of $x-$ and $y-$ values, and $S_x, S_y$ are the neutrosophic sample standard deviations.

Let's consider the example from the previous Table of Neutrosophic Sample of size 6.

$r_N$

$$= \frac{6 \cdot (215, 264) - (24, 26) \cdot (38, 44)}{\sqrt{6 \cdot (130, 152) - (24, 26)^2 \cdot [6 \cdot (362, 368) - (38, 44)^2]}}$$

$$= \frac{(6 \cdot 215, 6 \cdot 264) - (24 \cdot 38, 26 \cdot 44)}{\sqrt{[(6 \cdot 130, 6 \cdot 152) - (24^2, 26^2)] \cdot [(6 \cdot 362, 6 \cdot 468) - (38^2, 44^2)]}}$$

$$= \frac{(1290, 1584) - (912, 1144)}{\sqrt{[(780, 912) - (576, 676)] \cdot [(2172, 2808) - (1444, 1936)]}}$$

$$= \frac{(1290 - 1144, 1584 - 912)}{\sqrt{(780 - 676, 912 - 576) \cdot (2172 - 1936, 2808 - 1444)}}$$

$$= \frac{(146, 672)}{\sqrt{(104, 336) \cdot (236, 1364)}} = \frac{(146, 672)}{\sqrt{(104 \cdot 336, 336 \cdot 1364}}$$

$$= \frac{(146, 672)}{(\sqrt{34944}, \sqrt{458304})} \simeq \frac{(146, 672)}{(186.933, 676.982)}$$

$$= \left( \frac{146}{676.982}, \frac{672}{186.933} \right) \simeq (0.2157, 3.5949) \equiv (0.2157, 1].$$

In general $r_N$ is a subset of the interval $[-1, 1]$. If $r_N$ is a subset of $[0, 1]$ then the points $(x_i, y_i)$ for $i = 1, 2, \dots, n$, lie approximatively near a straight line of positive slope, while when $r_N$ is a subset centered or almost centered at 0 (or $r_N$ is nearly half in $[0, 1]$ and nearly half in $[-1, 0]$ then their is virtually no linear approximation but their may be a non-linear association between the points.

**Neutrosophic Random Numbers** is a sequence of numbers and indeterminacies occurring at random with equal probability.

The occurrence of a number or indeterminacy is not a guide to the numbers or indeterminacies that follow it, nor is it predicted from the numbers or indeterminacies that precede it.

Using eleven balls numbered 0 to 9 and another one ball that has its number erased (which one cannot read, that we note by $I$), then repeatedly withdrawing a ball and putting it back to the container.

We randomly generate the sequence:

$$2, 9, 9, I, 0, 7, 6, 2, 1, 1, I, 8 \ldots,$$

where $I$ = indeterminacy.

The computers can be enabled to generate neutrosophic random numbers using the same classical algorithms as for classical random numbers, but adding one or more states of indeterminacies with an equal chance of occurring each of them.

As a generalization we proposed the
**Neutrosophic Weighted Random Numbers**,
where each number $x_j$ has a different chance $p_j$ to occur, and each indeterminacy $I_j$ has a different chance $r_j$ to occur.

There are also cases when the numbers have to be in a given set; for example, each number should have $k$ digits.

# A Neutrosophic Normal Distribution

A Neutrosophic Normal Distribution of a continuous variable *X* is a classical normal distribution of *x*, but such that its mean μ or its standard deviation σ (or variance σ²), or both, are imprecise.

For example, μ, or σ, or both can be set(s) with two or more elements. The most common such distributions are when μ, σ, or both are intervals.

The neutrosophic frequency function formula is the same, except, as explained in the introduction, replacing μ by $μ_N$ and σ by $σ_N$:

$$X_N \sim N_N\left(μ_N, σ_N^2\right) = \frac{1}{σ_N\sqrt{2π}} exp\left(-\frac{(x - μ_N)^2}{2σ_N^2}\right),$$

where $X_N$ actually means that variable *X* may be neutrosophic (i.e. having some indeterminacy), and similarly $N_N(\cdot, \cdot)$ meaning that the normal distribution $N(\cdot, \cdot)$ may be neutrosophic (i.e. having some indeterminacy).

Instead of one bell-shaped curve, we may have two or more bell-shaped curves that have common and uncommon regions between them and are above the *x*-axis. Each one is symmetric with respect to the vertical line passing through the mean (x = μ).

As a **first neutrosophic example for normal distribution**, let's consider a normal distribution with μ = 15 and σ = [2, 3]. Thus the standard deviation is indeterminate.

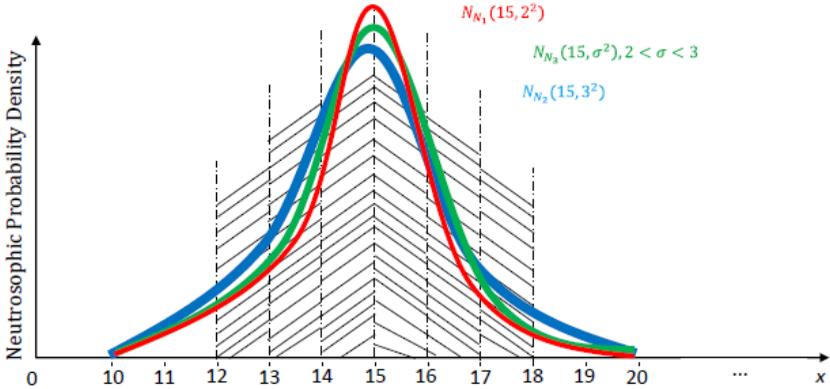

Within one standard deviation of the mean translates in this first example by:

$$\mu \pm \sigma = 15 \pm [2, 3] = [15 - 3, 15 + 3] = [12, 18],$$

or approximately 68% of values lie between

$$x \in [12, 18].$$

Within two standard deviations of the mean translates by:

$$\mu \pm 2\sigma = 15 \pm 2 \cdot [2, 3] = 15 \pm [4, 6] = [15 - 6, 15 + 6]$$
$$= [9, 21],$$

or approximately 95,4% of values lie between

$$x \in [9, 21].$$

We could also compute the last interval as:

$$[12, 18] \pm \sigma = [12, 18] \pm [2, 3] = [12 - 3, 18 + 3]$$
$$= [9, 21].$$

For three standard deviations:

$$\mu \pm 3\sigma = 15 \pm 3 \cdot [2,3] = 15 \pm [6,9] = [15 - 9, 15 + 9]$$
$$= [6, 24],$$

or we could compute it as

$$[9, 21] \pm [2,3] = [9 - 3, 21 + 3] = [6, 24],$$

and approximately 97,7% of values lie between

$$x \in [6, 24].$$

The area between the lowest and the highest curve for each portion represents the burden (indeterminacy) of the graph.

The neutrosophic normal distribution can be regarded as a bell-shape curve with heavy margins.

A random variable $X$ that has a neutrosophic normal distribution is called a *neutrosophic normalvariable.*

A **second neutrosophic examplefor normal distribution**where $\mu = [15, 17]$ and σ = 2, hence now μ is indeterminate.

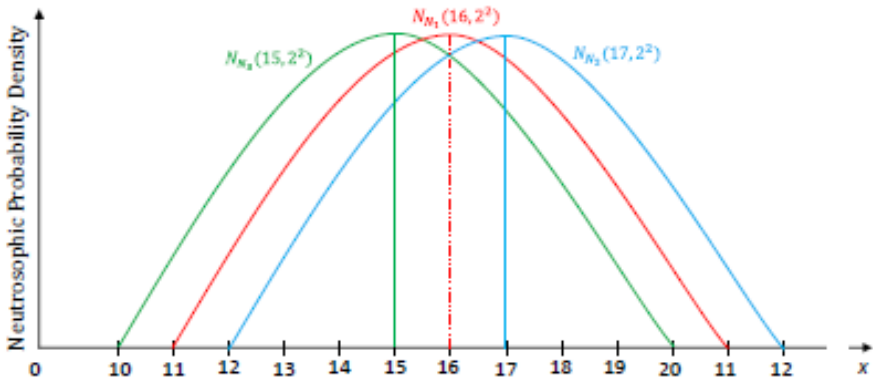

Similar discussion for the second example:

Within one standard deviation, i.e.

$\mu \pm \sigma = [15, 17] \pm 2 = [15 - 2, 17 + 2] = [13, 19]$,

approximately 68% of values lie between $x \in [13, 19]$.

Within two standard deviations, i.e.

$\mu \pm 2\sigma = [15, 17] \pm 2 \cdot 2 = [15, 17] \pm 4 = [15 - 4, 17 + 4]$
$$= [11, 21],$$

or computed as

$[13, 19] \pm \sigma = [13, 19] \pm 2 = [13 - 2, 19 + 2] = [11, 21]$.

And within three standard deviations, i.e.

$\mu \pm 3\sigma = [15, 17] \pm 3 \cdot 2 = [15, 17] \pm 6 = [15 - 6, 17 + 6]$
$$= [9, 23],$$

or computed as

$$[11, 21] \pm 2 = [11 - 2, 21 + 2] \pm 2 = [9, 23],$$

approximately 97.7% of values lie between
$$x \in [9, 23].$$

A **third neutrosophic example of normal distribution** with $\mu = [15, 17]$ and $\sigma = [2, 3]$, hence double indeterminacy, combines the previous second graph with the first one.

Of course, the vagueness becomes wider!

With $\mu = [15, 17]$ and $\sigma = [2, 3]$, we get:

Within one standard deviation of the mean, i.e.

$\mu \pm \sigma = [15, 17] \pm [2, 3] = [15 - 3, 17 + 3] = [12, 20]$,

approximately 68% of values lie between
$$x \in [12, 20].$$

Within two standard deviations of the mean, i.e.

$\mu \pm 2\sigma = [15, 17] \pm 2 \cdot [2, 3] = [15, 17] \pm [4, 6]$
$$= [15 - 6, 17 + 6] = [9, 23],$$

or computed as $[12, 20] \pm [2, 3] = [12 - 3, 20 + 3] = [9, 23]$,

approximately 95.4% of values lie between $x \in [9, 23]$.

And within three standard deviations of the mean, i.e.

$$\mu \pm 36 = [15, 17] \pm 3 \cdot [2, 3] = [15, 17] \pm [6, 9]$$
$$= [15 - 9, 17 + 9] = [6, 26],$$

or computed as $[9, 23] \pm [2, 3] = [9 - 3, 23 + 3] = [6, 26]$,

approximately 97.7% of values lie between
$$x \in [6, 26].$$

## Neutrosophication of Other Distributions.

In the same way, replacing one or more distribution parameters by a set, we can extend the classical distributions, such as: standard normal distribution, bivariate normal distribution, uniform distribution, sampling distribution, geometric distribution, hypergeometric distribution, Poisson distribution, chi-squared distribution, exponential distribution, frequency distribution, Pareto distribution, t-distribution, etc. to their corresponding neutrosophic versions.

The set replacing a crisp parameter may have two or more elements, or may be empty (the last case meaning that the parameter is unknown).

# A Neutrosophic Hypothesis

A Neutrosophic Hypothesis is a statement about the neutrosophic values of a single or several population characteristics.

The distinction between the classical (statistics) hypothesis and neutrosophic hypothesis is that in the neutrosophic statistics the variables that describe the population characteristics are neutrosophic (i.e. they have some indeterminate values, or several unknown values, or an inexact number of terms if the variable is discrete), or for the values that we compare at least one of the population characteristics is neutrosophic (i.e. indeterminate or unclear or vague value).

Similarly to the classical statistics, a **Neutrosophic Null Hypothesis**, denoted by $NH_0$, is the statement that is initially assumed to be true. While the **Neutrosophic Alternative Hypothesis**, denoted by $NH_a$, is the other hypothesis.

In carrying out a test of $NH_0$ versus $NH_a$ there are two possible conclusions: *reject $NH_0$* (if sample evidence suggest strongly that $NH_0$ is false), or *fail to reject $NH_0$* (if the sample does not support string evidence against $NH_0$).

**Examples**:
$$NH_0: \mu \in [90, 100]$$
$$NH_a: \mu < 90$$

$$NH_a: \mu > 100$$
$$NH_a: \mu \notin [90, 100],$$

where μ represents the classical average IQ of all children born since 1st January 2001.

$$NH_0: \pi = 0.2 \text{ or } 0.3$$
$$NH_a: \pi < 0.2$$
$$NH_a: \pi > 0.3$$
$$NH_a: \pi \in (0.2, 0.3)$$
$$NH_a: \mu \notin \{0.2, 0.3\},$$

where π represents the classical proportion of all Ford cars that need repair while under first year of warranty.

$$NH_0: p < 0.1 \text{ or } p > 0.9$$
$$NH_a: p = 0.1$$
$$NH_a: p = 0.9$$
$$NH_0: p > 0.1 \text{ and } p < 0.9$$
$$NH_a: p \in [0.1, 0.9],$$

where $p$ represents the classical proportion of outliers in a human population with respect to their height, i.e. percentage of people whose height is less than 150 cm, or percentage of people whose height is greater than 190 cm.

Neutrosophic Outliers are noticeably unusual values in the neutrosophic data; they can be crisp values or neutrosophic values.

$$NH_0: [\mu_{min}, \mu_{max}] > [0.45, 0.55],$$

which is equivalent to

$$\mu_{min} > 0.45$$

and

$$\mu_{max} > 0.55$$

where $\mu$ represents a neutrosophic percentage average of all electronic devices that get morally depreciated after three years from their fabrication; $[\mu_{min}, \mu_{max}]$ is a neutrosophic value (rough approximation).

$$NH_a: \mu_{min} = 0.45$$
$$NH_a: \mu_{max} = 0.55$$
$$NH_a: \mu_{min} < 0.45$$
$$NH_a: \mu_{max} < 0.55$$
$$NH_a: \mu_{min} < 0.45 \text{ or} \mu_{max} < 0.45.$$

$$NH_0: \mu = 7.0$$
$$NH_a: \mu < 7.0$$
$$NH_a: \mu > 7.0$$
$$NH_a: \mu \neq 7.0$$

A manufacturing plant made an approximate survey of its selling, survey done by two independent observes on different samples of same size. Their findings are close, yet different. The owner of the manufacturing plant decided to put both results together, taking for each period the [*min, max*] or [*inf, sup*] interval, in order to see the fluctuation of sales. The variable *x* that describes the survey is thus a neutrosophic one:

| Period | Sold Quantity (in thousands) |
|--------|------------------------------|
| 2001 | [4, 6] |
| 2002 | [7, 8] |
| 2003 | 5.5 or 6.0 |
| 2004 | (8.0, 8.8) |
| 2005 | 7.5 |

The null hypothesis that the average annual selling $\mu = 7.0$ is in the classical style, but the variable $x$ that $\mu$ is referring to is neutrosophic.

So we still have a neutrosophic hypothesis.

## Neutrosophic Hypothesis Testing Errors.

A census of a large population is hard or even impossible to due. That's why we have to use samples. The inference we are making from a neutrosophic sample characteristic to a population characteristic is subject to error.

Similarly to classical statistics, there are two types of errors:

1. *Neutrosophic Type I Error*, which is the error of rejecting $NH_0$ when $NH_0$ is true.

2. *Neutrosophic Type II Error*, which is the opposite of the previous error, i.e. the error of not rejecting $NH_0$ when $NH_0$ is false.

No matter what test we do, there is some chance that a neutrosophic type I error will be made, and there is some chance that a neutrosophic type II error will be made too.

For example, rejecting the hypothesis $H_0$: $\mu$ = 7.0 when it is true in one of the previous examples, would determine the owner of the manufacturing plant to take additional adjustments and spending money when not really needed.

While accepting $H_0$: $\mu$ = 7.0 when it is false, will damage the future selling.

Probabilities of neutrosophic type I error and type II error are denoted by $\alpha_N$ (*level of significance*) and respectively $\beta_N$.

Dealing with neutrosophic probabilities, $\alpha_N$ and $\beta_N$ can be subsets of the interval [0, 1]. The ideal test procedure would have $\alpha_N$ = $\beta_N \equiv$ 0, or $\alpha_N$ and $\beta_N$ as tiny intervals near zero.

For example, if $\alpha_N$ = [0.07, 0.10] in a test procedure, done with different samples, over and over, a true hypothesis $H_0$ is rejected about 7, 8, 9, or 10 times in a hundred.

If $\beta_N$ = [0.07, 0.10], then a false hypothesis $H_0$ is accepted about 7-10 times in a hundred.

**Example**.

A car manufacturer pretends that between 80% and 90% of its car need no repair during the first 2 years of driving. In order to check the claim, a consumer agency obtains a random sample of 50 purchasers and investigate them whether or not their cars needed repair during the first 2 years of driving. Let $p$ denote the sample proportion of responses that indicate no repair, and let π denote the true proportion of no repairs (called successes). The appropriate neutrosophic hypotheses are:

$$NH_0 : \pi \in [0.8, 0.9] \text{ versus } NH_a : \pi < 0.8$$

in order to check if the sample evidence suggests that $\pi < 0.9$.

Neutrosophic Type I Error is to consider the car manufacturer's claim fallacious (i.e. $\pi < 0.8$) while in fact it is correct.

And Neutrosophic Type II Error if the consumer agency fails to detect the manufacturer's incorrect claim.

For avoiding serious consequences the consumer agency decides a type I error probability of $[0.01, 0.05]$ but no larger can be tolerated. So

$\alpha = [0.01, 0.05]$ is used for developing a test procedure.

We recall, from classical statistics, that a *classical standard normal distribution of a random variable z*, is a normal distribution with the mean value

$$\mu = 0$$

and standard deviation

$$\sigma = 1.$$

Its corresponding curve is called **standard normal curve** or *z curve*.

A **z critical value** captures the lower-tail or upper-tail area, or the central area under the $z$ curve.

The table of the most used z critical values in classical statistics:

| Critical value, z | Area to the right of z | Area to the left of –z | Area between –z and z |
|---|---|---|---|
| 1.28 | .10 | .10 | .80 |
| 1.645 | .05 | .05 | .90 |
| 1.96 | .025 | .025 | .95 |
| 2.33 | .01 | .01 | .98 |
| 2.58 | .005 | .005 | .99 |
| 3.09 | .001 | .001 | .998 |
| 3.29 | .0005 | .0005 | .999 |

A *normally distributed random variable x can bestandardized* as

$$z = \frac{x - \mu}{\sigma},$$

where $\mu = x's$ mean value,

and $\sigma = x's$ standard distribution.

If the **neutrosophic null hypothesis** about variable $x$ is:

$$NH_0 : \mu \in [a, b],$$

where $[a, b]$, with $a \leq b$, is the hypothesized interval, then the **neutrosophic test statistic** is:

$$z = \frac{\bar{x} - [a.b]}{s/\sqrt{n}}$$

where      $\bar{x}$ is the sample mean,

$s$ is the sample standard deviation,

and      $n$ is the sample size, with $n > 30$.

Variable $z$ has approximately a neutrosophic standard normal distribution.

In neutrosophic statitics, $\bar{x}$, $s$ and even $n$ can be sets (not necessarily crisp numbers).

## Alternative Hypotheses.

$H_a: \mu > b$; Reject $H_0$ if $\min z > z$ critical value (upper-tailed test);

$H_a: \mu < a$; Reject $H_0$ if $\max z < -z$ critical value (lower-tailed test);

$H_a: \mu \notin [a, b]$; Reject $H_0$ if: either $\min z > z$ critical value, or $\max z < -z$ critical value (two-tailed test).

## Example.

Let's consider the exam-anxiety scores for a sample of an American College students were the following:

$n = 64, \bar{x} = [48.0, 50.0]$, and $s = 25$.

Then $\mu = $ true mean exam-anxiety.

$$H_0: \mu \in [40.0, 41.0]$$
$$H_a: \mu > 41.0.$$

The neutrosophic test statistics is:

$$z = \frac{[48.0, 50.0] - [40.0, 41.0]}{25/\sqrt{64}} = \frac{[48.0 - 41.0, 50.0 - 40.0]}{25/8}$$

$$= \frac{[7.0, 10.0]}{25/8} = \frac{8 \cdot [7.0, 10.0]}{25} = \frac{[56.0, 80.0]}{25}$$

$$= \left[\frac{56.0}{25}, \frac{80.0}{25}\right] = [2.24, 3.20].$$

For $\alpha = 0.10$ the corresponding one-tailed $z$ critical value from the previous table is 1.28. Hence $H_0$ is rejected because $z = [2.24, 3.20] > 1.28$.

In conclusion, the mean exam-anxiety score is higher than 41.0.

# The Neutrosophic Level of Significance

**The Neutrosophic Level of Significance α** may be a set, not necessarily a crisp number as in classical statistics.

For example, $\alpha_4 = [0.01, 0.10]$ is a neutrosophic level of significance α, where α varies in the interval $[0.01, 0.10]$.

A **Neutrosophic P-Value** is defined in the same way as in classical statistics: the smallest level of significance at which a null hypothesis $H_0$ can be rejected.

The distinction between classical P-value and neutrosophic P-value is that the neutrosophic P-value is not a crisp number as in classical statistics, but a set (in many applications it is an interval).

Neutrosophic P-Value $= P(z > z$critical value, when $H_0$ is true), where $P(\cdot)$ means classical probability calculated assuming that $H_0$ is true, probability of observing a test statistic value being more extreme than is was actually obtained.

Suppose one has calculated the neutrosophic P-value at the particular level of significance α, where *a is a crisp positive number.*

1. If $max\{neutrosophic P - value\} \leq \alpha$, then reject $H_0$ at level α.

2. If $min\{neutrosophic P - value\} > \alpha$, then do not reject $H_0$ at level α.

3. If $min\{neutrosophicP - value\} < \alpha < max\{neutrosophicP - value\}$ then there is an indeterminacy. Thus

$$\frac{\alpha - min\{neutrosophicP - value\}}{max\{neutrosophicP - value\} - min\{neutrosophicP - value\}}$$

is the chance of rejecting $H_0$ at level ɑ,
and

$$\frac{max\{neutrosophicP - value\} - \alpha}{max\{neutrosophicP - value\} - min\{neutrosophicP - value\}}$$

is the chance of not rejecting $H_0$ at level ɑ.

Let $\alpha_N$ be a set.

4. If $max\{neutrosophicP - value\} \leq min\{\alpha_N\}$, then reject $H_0$ at level $\alpha_N$.

5. If $min\{neutrosophicP - value\} > max\{\alpha_N\}$, then do not reject $H_0$ at level $\alpha_N$.

6. If the two sets, those of the neutrosophic P-value and of the neutrosophic level of significance $\alpha_N$ intersect, one has indeterminacy. And one can compute the chance of rejecting $H_0$ at level $\alpha_N$, and the chance of not rejecting $H_0$ at level $\alpha_N$.

In classical statistics, the *P-value* is computed considering the *Table of Standard Normal Probabilities.*

a. P-value is the area under the $z$ curve to the right of computed $z$, for *Upper-tailed z test.*

b. P-value is the area under the $z$-curve to the left of computed $z$, for *Lower-tailed z test.*

c. P-value is twice the area captured in the tail corresponding to the computed $z$, for *Two-tailed z test.*

Let's insert from the classical statistics the *Standard Normal Cumulative Probability Table* [for positive z-values only, since this is needed in our below example]*:*

**Standard Normal Cumulative Probability Table**

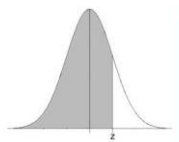

Cumulative probabilities for POSITIVE z-values are shown in the following table:

| z | 0.00 | 0.01 | 0.02 | 0.03 | 0.04 | 0.05 | 0.06 | 0.07 | 0.08 | 0.09 |
|---|------|------|------|------|------|------|------|------|------|------|
| 0.0 | 0.5000 | 0.5040 | 0.5080 | 0.5120 | 0.5160 | 0.5199 | 0.5239 | 0.5279 | 0.5319 | 0.5359 |
| 0.1 | 0.5398 | 0.5438 | 0.5478 | 0.5517 | 0.5557 | 0.5596 | 0.5636 | 0.5675 | 0.5714 | 0.5753 |
| 0.2 | 0.5793 | 0.5832 | 0.5871 | 0.5910 | 0.5948 | 0.5987 | 0.6026 | 0.6064 | 0.6103 | 0.6141 |
| 0.3 | 0.6179 | 0.6217 | 0.6255 | 0.6293 | 0.6331 | 0.6368 | 0.6406 | 0.6443 | 0.6480 | 0.6517 |
| 0.4 | 0.6554 | 0.6591 | 0.6628 | 0.6664 | 0.6700 | 0.6736 | 0.6772 | 0.6808 | 0.6844 | 0.6879 |
| 0.5 | 0.6915 | 0.6950 | 0.6985 | 0.7019 | 0.7054 | 0.7088 | 0.7123 | 0.7157 | 0.7190 | 0.7224 |
| 0.6 | 0.7257 | 0.7291 | 0.7324 | 0.7357 | 0.7389 | 0.7422 | 0.7454 | 0.7486 | 0.7517 | 0.7549 |
| 0.7 | 0.7580 | 0.7611 | 0.7642 | 0.7673 | 0.7704 | 0.7734 | 0.7764 | 0.7794 | 0.7823 | 0.7852 |
| 0.8 | 0.7881 | 0.7910 | 0.7939 | 0.7967 | 0.7995 | 0.8023 | 0.8051 | 0.8078 | 0.8106 | 0.8133 |
| 0.9 | 0.8159 | 0.8186 | 0.8212 | 0.8238 | 0.8264 | 0.8289 | 0.8315 | 0.8340 | 0.8365 | 0.8389 |
| 1.0 | 0.8413 | 0.8438 | 0.8461 | 0.8485 | 0.8508 | 0.8531 | 0.8554 | 0.8577 | 0.8599 | 0.8621 |
| 1.1 | 0.8643 | 0.8665 | 0.8686 | 0.8708 | 0.8729 | 0.8749 | 0.8770 | 0.8790 | 0.8810 | 0.8830 |
| 1.2 | 0.8849 | 0.8869 | 0.8888 | 0.8907 | 0.8925 | 0.8944 | 0.8962 | 0.8980 | 0.8997 | 0.9015 |
| 1.3 | 0.9032 | 0.9049 | 0.9066 | 0.9082 | 0.9099 | 0.9115 | 0.9131 | 0.9147 | 0.9162 | 0.9177 |
| 1.4 | 0.9192 | 0.9207 | 0.9222 | 0.9236 | 0.9251 | 0.9265 | 0.9279 | 0.9292 | 0.9306 | 0.9319 |
| 1.5 | 0.9332 | 0.9345 | 0.9357 | 0.9370 | 0.9382 | 0.9394 | 0.9406 | 0.9418 | 0.9429 | 0.9441 |
| 1.6 | 0.9452 | 0.9463 | 0.9474 | 0.9484 | 0.9495 | 0.9505 | 0.9515 | 0.9525 | 0.9535 | 0.9545 |
| 1.7 | 0.9554 | 0.9564 | 0.9573 | 0.9582 | 0.9591 | 0.9599 | 0.9608 | 0.9616 | 0.9625 | 0.9633 |
| 1.8 | 0.9641 | 0.9649 | 0.9656 | 0.9664 | 0.9671 | 0.9678 | 0.9686 | 0.9693 | 0.9699 | 0.9706 |
| 1.9 | 0.9713 | 0.9719 | 0.9726 | 0.9732 | 0.9738 | 0.9744 | 0.9750 | 0.9756 | 0.9761 | 0.9767 |
| 2.0 | 0.9772 | 0.9778 | 0.9783 | 0.9788 | 0.9793 | 0.9798 | 0.9803 | 0.9808 | 0.9812 | 0.9817 |
| 2.1 | 0.9821 | 0.9826 | 0.9830 | 0.9834 | 0.9838 | 0.9842 | 0.9846 | 0.9850 | 0.9854 | 0.9857 |
| 2.2 | 0.9861 | 0.9864 | 0.9868 | 0.9871 | 0.9875 | 0.9878 | 0.9881 | 0.9884 | 0.9887 | 0.9890 |
| 2.3 | 0.9893 | 0.9896 | 0.9898 | 0.9901 | 0.9904 | 0.9906 | 0.9909 | 0.9911 | 0.9913 | 0.9916 |
| 2.4 | 0.9918 | 0.9920 | 0.9922 | 0.9925 | 0.9927 | 0.9929 | 0.9931 | 0.9932 | 0.9934 | 0.9936 |
| 2.5 | 0.9938 | 0.9940 | 0.9941 | 0.9943 | 0.9945 | 0.9946 | 0.9948 | 0.9949 | 0.9951 | 0.9952 |
| 2.6 | 0.9953 | 0.9955 | 0.9956 | 0.9957 | 0.9959 | 0.9960 | 0.9961 | 0.9962 | 0.9963 | 0.9964 |
| 2.7 | 0.9965 | 0.9966 | 0.9967 | 0.9968 | 0.9969 | 0.9970 | 0.9971 | 0.9972 | 0.9973 | 0.9974 |
| 2.8 | 0.9974 | 0.9975 | 0.9976 | 0.9977 | 0.9977 | 0.9978 | 0.9979 | 0.9979 | 0.9980 | 0.9981 |
| 2.9 | 0.9981 | 0.9982 | 0.9982 | 0.9983 | 0.9984 | 0.9984 | 0.9985 | 0.9985 | 0.9986 | 0.9986 |
| 3.0 | 0.9987 | 0.9987 | 0.9987 | 0.9988 | 0.9988 | 0.9989 | 0.9989 | 0.9989 | 0.9990 | 0.9990 |
| 3.1 | 0.9990 | 0.9991 | 0.9991 | 0.9991 | 0.9992 | 0.9992 | 0.9992 | 0.9992 | 0.9993 | 0.9993 |
| 3.2 | 0.9993 | 0.9993 | 0.9994 | 0.9994 | 0.9994 | 0.9994 | 0.9994 | 0.9995 | 0.9995 | 0.9995 |
| 3.3 | 0.9995 | 0.9995 | 0.9995 | 0.9996 | 0.9996 | 0.9996 | 0.9996 | 0.9996 | 0.9996 | 0.9997 |
| 3.4 | 0.9997 | 0.9997 | 0.9997 | 0.9997 | 0.9997 | 0.9997 | 0.9997 | 0.9997 | 0.9997 | 0.9998 |

In the previous example,

$H_0 \cdot \mu \, \epsilon [40.0, 41.0]$ versus $H_a: \mu > 41.0$,

we found the neutrosophic $z = [2.24, 3.20]$. We have an Upper-tailed $z$ test.

From the above Table of Standard Normal Probabilities, the area under the $z$ curve to the right of $z_1 = 2.24$ is $1 - 0.9875 = 0.0125$

while for

$z_2 = 3.20$ is $1 - 0.9993 = 0.0007$.

Thus, the neutrosophic

$$P - value = [0.0007, 0.0125].$$

At the level of significance $\alpha_1 = 0.10$, reject $H_0$ since

$$max[0.0007, 0.0125] = 0.0125 < 0.10.$$

At the level of significance $\alpha_2 = 0.0005$, do not reject $H_0$ since

$$max[0.0007, 0.0125] = 0.0125 > 0.0005.$$

At the level of significance $\alpha_3 = 0.01$, one has indeterminacy since

$0.01 \in [0.0007, 0.0125]$; therefore:

chance of rejecting $H_0$ at level $\alpha_3 = 0.01$ is

$$\frac{0.01 - 0.0007}{0.0125 - 0.0007} = \frac{0.0093}{0.0118} \simeq 79\%$$

and chance of not rejecting $H_0$ at level $\alpha_3 = 0.01$ is

$$\frac{0.0125 - 0.01}{0.0125 - 0.07} = \frac{0.0025}{0.0118} \simeq 21\%.$$

# The Neutrosophic Confidence Interval

**The Neutrosophic Confidence Interval** for a population characteristics is defined, similarly to the classical statistics, as an interval of plausible neutrosophic values of the characteristic.

The neutrosophic value of the characteristic is captured inside the interval with a chosen degree of confidence.

A **confidence level** is associated with each neutrosophic confidence interval, as in classical statistics. It tells us how much confidence we have in procedure used in constructing the neutrosophic confidence interval.

The classical formulas for the confidence interval are extended from crisp variables to neutrosophic variables (i.e. variables whose values are sets):

1. When the neutrosophic value of the population standard deviation σ is known, the **Large-Sample Neutrosophic Confidence Interval for the Population Mean μ** is:

$$\bar{x} \pm (z\text{critical value}) \cdot \frac{\sigma}{\sqrt{n}}$$

where $\bar{x}$ is the large-sample neutrosophic mean, and $n$ is the neutrosophic size of the large-sample.

Therefore $\bar{x}$, σ, and/or $n$ may be sets instead of crisp numbers.

2. When the neutrosophic value of the population standard deviation σ is unknown (as in most practical applications), and the sample size exceeds 30, one uses the sample standard deviation *s* instead of σ for computing the **Neutrosophic Confidence Interval for the Population Mean μ**:

$$\bar{x} \pm (z\text{critical value}) \cdot \frac{s}{\sqrt{n}}.$$

For both formulas, the *z* critical value 1.645 corresponds to the confidence level of 90%, the *z* critical value 1.96 corresponds to the confidence level of 95%, and the *z* critical value 2,58 corresponds to the confidence level of 99%, similarly as in classical statistics.

The confidence level of, for example, 90% does not refer to the chance that the population mean μ is captured in an interval, but to the percentage of all possible successful samples (i.e. samples for which μ is included in the confidence interval).

An **Example**.

Many individuals partially loose vision because of exposure to dust.

On a study involving 60 people (a sample), that were constantly exposed to dust to their construction work places, in average they lost 18%-20% of their vision accuracy, with a sample standard deviation of 4%-5%.

The study investigator wishes a 90% confidence interval for μ. Hence:

$$\bar{x} = [18, 20]$$
$$z\text{critical value} = 1.645$$

$$s = [4, 5]$$
$$n = 60.$$

Therefore, the neutrosophic confidence interval for the population mean µ is:

$$[18, 20] \pm (1.645) \cdot \frac{[4, 5]}{\sqrt{60}}$$

$$= [18, 20] \pm \left[ \frac{1.645(4)}{\sqrt{60}}, \frac{1.645(5)}{\sqrt{60}} \right]$$

$$\backsimeq [18, 20] \pm [0.85, 1.06].$$

Let's split into two parts:

$$[18, 20] + [0.85, 1.06] = [18 + 0.85, 20 + 1.06]$$
$$= [18.85, 21.06],$$

and

$$[18, 20] - [0.85, 1.06] = [18 - 1.06, 20 - 0.85]$$
$$= [16.94, 19.15].$$

Combining these two cases we get the neutrosophic confidence interval:

$$[16.94, 21.06].$$

The Neutrosophic Sample Size to estimate, within the amount B, with $c\%$ confidence, of the population mean µ is:

$$n_N = \left[ \frac{(z \text{critical value}) \cdot \sigma}{B} \right],$$

where $z$ critical value should correspond to the $c\%$ confidence,

σ is the population standard variation,

and $n_N$ is the resulting neutrosophic sample size, hence $n_N$ may be a set (especially an interval).

For surety, we can take the sample size as $[max\{n_N\}]$, where $\lceil \; \rceil$ means superior integer part.

Let's see **an Example**.

The business department wishes to estimate the annual cost of office supplies for faculty at the University of New Mexico to be within $40 of the true population mean. The business department wants a 95% confidence in their result accuracy.

How large should the sample be?

Because σ is not known, it can be approximated as

$$\sigma \approx \frac{\text{range}}{4}$$

as in classical statistics.

Range is the difference between the highest and lowest costs.

The amount spent on office supplies varied between $500-$550 to $100-$150. Then

$$\sigma \approx \frac{[500, 550] - [100, 150]}{4} = \frac{[500 - 150, 550 - 100]}{4}$$
$$= \frac{[350, 450]}{4} = \left[\frac{350}{4}, \frac{450}{4}\right]$$
$$= [87.50, 137.50].$$

Further, $B = 40$, $z$ critical value is 1.96, and:

$$n_N = \left[\frac{1.96[87.50, 137.50]}{40}\right]^2 = \left[\frac{1.96(87.50)}{40}, \frac{1.96(137.50)}{40}\right]^2$$
$$= [4.2875, 6.7375]^2 = [4.2875^2, 6.7375^2]$$
$$\simeq [18.38, 45.39].$$

Now

$$[\max[18.38, 45.39]] = [45.39] = 46.$$

Therefore the sample size should be 46.

# Large-Sample Neutrosophic Confidence Interval for the Population Proportion

Using the classical statistics one can define (in the same way) the **Large-Sample Neutrosophic Confidence Interval for the Population Proportion π**:

$$p \pm (z\text{critical value}) \cdot \sqrt{\frac{p(1-p)}{n}}$$

for the case when $\min\{np\} \geq 5$ and $\min\{n \cdot (1 - p)\} \geq 5$,

where

p = sample proportion = number of sample individuals that possess the property of interest divided by sample's size;

n = sample's size;

π = population proportion = $\frac{\text{number of population individuals that possess the property of interest}}{\text{total number of population individuals}}$,

with the distinction from the classical statistics that in neutrosophic statistics the parameters p and n may be setsinstead of crisp numbers, and the z critical value may be a set as well(for example it may be [1.645, 1.96], i.e. confidence level of [90, 95]%).

The neutrosophic sample statistics p, for $min\{n\}$ large enough, has a neutrosophic sampling distribution (normal curve) that approximates the population mean $\pi$ and its standard deviation $\sqrt{\frac{\pi(1-\pi)}{n}}$.

Let's see an **Example**.

A survey on a sample of $200 - 220$ consumers is done at a car dealer asking the following question: "Would you be willing to trade in your old car when buying a new car?" The number of yes's was 150. The confidence level should be 99%. If $\pi$ denotes the proportion of all consumers who would trade in their old cars, one may consider $p$ a point estimate for $\pi$:

$$p = \frac{150}{\{200,\ 201, \dots, 220\}} \simeq \left[\frac{150}{220}, \frac{150}{200}\right] \simeq [0.68, 0.75].$$

The sample's size $\{200, 201, \dots, 220\}$ means that the surveyer was not sure about 20 people if they were or not custumers of this car dealer. So, the sample's size is indeterminate (approximated by the set $\{200, 201, \dots, 220\}$),

$z$ critical value = 2.58.

$$min\{np\} = min\{\{200, 201, \dots, 220\} \cdot [0.68, 0.75]\}$$
$$= 200(0.68) = 136 > 5;$$

$$min\{n(1-p)\} = min\{\{200, 201, \dots 220\}$$
$$\cdot (1 - [0.68, 0.75])\}$$
$$= 200 \cdot min([1 - 0.75, 1 - 0.68])$$
$$= 200 \cdot min([0.25, 0.32]) = 200(0.25)$$
$$= 50 > 5.$$

The large-sample neutrosophic confidence interval for π is:

$$[0.68, 0.75] \pm 2.58 \cdot \sqrt{\frac{[0.68, 0.75] \cdot (1 - [0.68, 0.75])}{\{200, 201, \dots, 220\}}}$$
$$= [0.68, 0.75] \pm 2.58$$
$$\cdot \sqrt{\frac{[0.68, 0.75] \cdot [0.25, 0.32]}{\{200, 201, \dots, 220\}}}$$
$$= [0.68, 0.75] \pm 2.58$$
$$\cdot \sqrt{\left[\frac{0.68(0.25)}{220}, \frac{0.75(0.32)}{200}\right]}$$
$$= [0.68, 0.75] \pm 2.58$$
$$\cdot \sqrt{[0.000773, 0.001200]}$$
$$= [0.68, 0.75] \pm 2.58$$
$$\cdot \sqrt{0.000773}, \sqrt{0.001200}$$
$$= [0.68, 0.75] \pm 2.58$$
$$\cdot [0.027803, 0.034641]$$
$$= [0.68, 0.75] \pm [0.071732, 0.089374].$$

Split it into two parts:

$$[0.68, 0.75] + [0.71732, 0.089374]$$
$$= [0.751732, 0.839374],$$

and

$[0.68, 0.75] − [0.071732, 0.089374]$
$$= [0.68 − 0.089374, 0.75 − 0.071732]$$
$$= [0.590626, 0.678268].$$

Combining both results in a conservative mode, we get:

$$[0.590626, 0.839374].$$

The formula for choosing the neutrosophic sample size is the same as in classical statistics, but using sets instead of crisp numbers:

$$n = \pi(1 − \pi) \cdot \left[ \frac{z\text{critical value}}{B} \right]^2$$

where B = the specific error bound.

If π cannot be estimated using prior neutrosophic information, one uses π = 0.5 which gives a conservatively large sample value (i.e. a larger n than any other value of π would do).

# The Neutrosophic Central Limit Theorem

The Neutrosophic Central Limit Theorem, which is an extension of the classical Central Limit Theorem, can be safely applied if $min\{n\}$ exceeds 30, where $n$ is the neutrosophic sample size (i.e. $n$ may be a set).

The Neutrosophic Central Limit Theorem states that the neutrosophic sampling distribution of $\bar{x}$ si approximated by a neutrosophic normal curve when $min\{n\}$ is sufficiently large, no matter how is the population distribution.

Of course, if the population distribution is normal, then $min\{n\}$ may be smaller than 30, and the neutrosophic sampling distribution of $\bar{x}$ is normal too for any neutrosophic sample size $n$. But, if the population distribution is not normal, then $min\{n\}$ should be greater than 30, and the neutrosophic sampling distribution of $\bar{x}$ is only an approximation to the normal curve: the larger is $min\{n\}$, the better approach.

The last result has enabled the *neutrosophic statisticians* in order to infer a population mean, to develop large sample neutrosophic procedures even when one deals with an unknown shape of the population distribution.

Using similar notations:

$n$ = random neutrosophic sample size;

$\bar{x}$ = neutrosophic mean of the sample size;

$\mu$ = population mean;

$\sigma$ = population standard deviation;

$\mu_{\bar{x}}$ = neutrosophic mean of the $\bar{x}$ distribution; and

$\sigma_{\bar{x}}$ = neutrosophic standard deviation of the $\bar{x}$ distribution;

one has, as in classical statistics:

$$\mu_{\bar{x}} = \mu,$$
$$\text{and } \sigma_{\bar{x}} = \frac{\sigma}{\sqrt{n}}.$$

The neutrosophic central limit theorem does not apply, as in classical statistics, when $min\{n\}$ is small and the shape of the population distribution is unknown.

Let's introduce the **Small-Sample Neutrosophic $t$ Confidence Interval for the Mean of the Normal Population**, which is just a neutrosophication of the classical one-sample $t$ confidence interval for the population mean μ:

$$\bar{x} \pm (t\text{critical values}) \cdot \frac{s}{\sqrt{n}}$$

where similarly:

$\bar{x}$ = neutrosophic sample mean;

$s$ = neutrosophic sample standard deviation;

$n$ = neutrosophic sample size;

and

$t$critical value is based on

$min\{n\} - 1$ degrees of freedom $(df)$.

$\bar{x}, s,$ and $n$ may be sets instead of crisp numbers.

For small $min\{n\}$, the neutrosophic $t$ confidence interval for the population mean μ is appropriate when the population distribution is normal or approximately normal. Otherwise, another method should be employed.

The **neutrosophic t distribution** is more spread out, of course, than the neutrosophic standard normal $(z)$ curve, because the use of $s$, instead of population deviation σ, produces extra variability.

The neutrosophic $t$ distributions are distinguished from one another by the **degree of freedom**, which can be a positive integer greater than or equal to 1, or a set of positive integers greater than or equal to 1, for example:

$$\{n, n + 1, ..., n + m\}.$$

The higher is $min\{n\}$, the closer the neutrosophic $t$ distribution is to the neutrosophic $z$ curve. For $min\{n\} > 120$ one may use the $z$ critical values. The neutrosophic $t$ curve, for a fixed number of degrees of freedom, is in general bell-shaped and centered at zero in neutrosophic style way.

An **Example**.

A small random sample of 18 workers, at the Rail Road, was investigated regarding the weights these workers are able to lift in their work place. The neutrosophic sample average found was $\bar{x}$ between 8 kg and 10 kg, with a standard deviation $s$ between 3-4 kg.

Let's say a confidence level of 95% is required for capturing the population mean μ.Thus:

$$\bar{x} = [8, 10](\text{an interval})$$
$$s = [3,4](\text{an interval})$$
$$n = 18,$$

hence a small sample size, which requires a neutrosophic $t$ critical value based on $18 - 1 = 17\, df$.

From the below classical statistics*Table of t Critical Values*,we find out that for 95% confidence level and 17 df, the corresponding

$$t\text{critical value} = 2.11.$$

## t Table

| cum. prob | $t_{.50}$ | $t_{.75}$ | $t_{.80}$ | $t_{.85}$ | $t_{.90}$ | $t_{.95}$ | $t_{.975}$ | $t_{.99}$ | $t_{.995}$ | $t_{.999}$ | $t_{.9995}$ |
|---|---|---|---|---|---|---|---|---|---|---|---|
| one-tail | 0.50 | 0.25 | 0.20 | 0.15 | 0.10 | 0.05 | 0.025 | 0.01 | 0.005 | 0.001 | 0.0005 |
| two-tails | 1.00 | 0.50 | 0.40 | 0.30 | 0.20 | 0.10 | 0.05 | 0.02 | 0.01 | 0.002 | 0.001 |
| df | | | | | | | | | | | |
| 1 | 0.000 | 1.000 | 1.376 | 1.963 | 3.078 | 6.314 | 12.71 | 31.82 | 63.66 | 318.31 | 636.62 |
| 2 | 0.000 | 0.816 | 1.061 | 1.386 | 1.886 | 2.920 | 4.303 | 6.965 | 9.925 | 22.327 | 31.599 |
| 3 | 0.000 | 0.765 | 0.978 | 1.250 | 1.638 | 2.353 | 3.182 | 4.541 | 5.841 | 10.215 | 12.924 |
| 4 | 0.000 | 0.741 | 0.941 | 1.190 | 1.533 | 2.132 | 2.776 | 3.747 | 4.604 | 7.173 | 8.610 |
| 5 | 0.000 | 0.727 | 0.920 | 1.156 | 1.476 | 2.015 | 2.571 | 3.365 | 4.032 | 5.893 | 6.869 |
| 6 | 0.000 | 0.718 | 0.906 | 1.134 | 1.440 | 1.943 | 2.447 | 3.143 | 3.707 | 5.208 | 5.959 |
| 7 | 0.000 | 0.711 | 0.896 | 1.119 | 1.415 | 1.895 | 2.365 | 2.998 | 3.499 | 4.785 | 5.408 |
| 8 | 0.000 | 0.706 | 0.889 | 1.108 | 1.397 | 1.860 | 2.306 | 2.896 | 3.355 | 4.501 | 5.041 |
| 9 | 0.000 | 0.703 | 0.883 | 1.100 | 1.383 | 1.833 | 2.262 | 2.821 | 3.250 | 4.297 | 4.781 |
| 10 | 0.000 | 0.700 | 0.879 | 1.093 | 1.372 | 1.812 | 2.228 | 2.764 | 3.169 | 4.144 | 4.587 |
| 11 | 0.000 | 0.697 | 0.876 | 1.088 | 1.363 | 1.796 | 2.201 | 2.718 | 3.106 | 4.025 | 4.437 |
| 12 | 0.000 | 0.695 | 0.873 | 1.083 | 1.356 | 1.782 | 2.179 | 2.681 | 3.055 | 3.930 | 4.318 |
| 13 | 0.000 | 0.694 | 0.870 | 1.079 | 1.350 | 1.771 | 2.160 | 2.650 | 3.012 | 3.852 | 4.221 |
| 14 | 0.000 | 0.692 | 0.868 | 1.076 | 1.345 | 1.761 | 2.145 | 2.624 | 2.977 | 3.787 | 4.140 |
| 15 | 0.000 | 0.691 | 0.866 | 1.074 | 1.341 | 1.753 | 2.131 | 2.602 | 2.947 | 3.733 | 4.073 |
| 16 | 0.000 | 0.690 | 0.865 | 1.071 | 1.337 | 1.746 | 2.120 | 2.583 | 2.921 | 3.686 | 4.015 |
| 17 | 0.000 | 0.689 | 0.863 | 1.069 | 1.333 | 1.740 | 2.110 | 2.567 | 2.898 | 3.646 | 3.965 |
| 18 | 0.000 | 0.688 | 0.862 | 1.067 | 1.330 | 1.734 | 2.101 | 2.552 | 2.878 | 3.610 | 3.922 |
| 19 | 0.000 | 0.688 | 0.861 | 1.066 | 1.328 | 1.729 | 2.093 | 2.539 | 2.861 | 3.579 | 3.883 |
| 20 | 0.000 | 0.687 | 0.860 | 1.064 | 1.325 | 1.725 | 2.086 | 2.528 | 2.845 | 3.552 | 3.850 |
| 21 | 0.000 | 0.686 | 0.859 | 1.063 | 1.323 | 1.721 | 2.080 | 2.518 | 2.831 | 3.527 | 3.819 |
| 22 | 0.000 | 0.686 | 0.858 | 1.061 | 1.321 | 1.717 | 2.074 | 2.508 | 2.819 | 3.505 | 3.792 |
| 23 | 0.000 | 0.685 | 0.858 | 1.060 | 1.319 | 1.714 | 2.069 | 2.500 | 2.807 | 3.485 | 3.768 |
| 24 | 0.000 | 0.685 | 0.857 | 1.059 | 1.318 | 1.711 | 2.064 | 2.492 | 2.797 | 3.467 | 3.745 |
| 25 | 0.000 | 0.684 | 0.856 | 1.058 | 1.316 | 1.708 | 2.060 | 2.485 | 2.787 | 3.450 | 3.725 |
| 26 | 0.000 | 0.684 | 0.856 | 1.058 | 1.315 | 1.706 | 2.056 | 2.479 | 2.779 | 3.435 | 3.707 |
| 27 | 0.000 | 0.684 | 0.855 | 1.057 | 1.314 | 1.703 | 2.052 | 2.473 | 2.771 | 3.421 | 3.690 |
| 28 | 0.000 | 0.683 | 0.855 | 1.056 | 1.313 | 1.701 | 2.048 | 2.467 | 2.763 | 3.408 | 3.674 |
| 29 | 0.000 | 0.683 | 0.854 | 1.055 | 1.311 | 1.699 | 2.045 | 2.462 | 2.756 | 3.396 | 3.659 |
| 30 | 0.000 | 0.683 | 0.854 | 1.055 | 1.310 | 1.697 | 2.042 | 2.457 | 2.750 | 3.385 | 3.646 |
| 40 | 0.000 | 0.681 | 0.851 | 1.050 | 1.303 | 1.684 | 2.021 | 2.423 | 2.704 | 3.307 | 3.551 |
| 60 | 0.000 | 0.679 | 0.848 | 1.045 | 1.296 | 1.671 | 2.000 | 2.390 | 2.660 | 3.232 | 3.460 |
| 80 | 0.000 | 0.678 | 0.846 | 1.043 | 1.292 | 1.664 | 1.990 | 2.374 | 2.639 | 3.195 | 3.416 |
| 100 | 0.000 | 0.677 | 0.845 | 1.042 | 1.290 | 1.660 | 1.984 | 2.364 | 2.626 | 3.174 | 3.390 |
| 1000 | 0.000 | 0.675 | 0.842 | 1.037 | 1.282 | 1.646 | 1.962 | 2.330 | 2.581 | 3.098 | 3.300 |
| z | 0.000 | 0.674 | 0.842 | 1.036 | 1.282 | 1.645 | 1.960 | 2.326 | 2.576 | 3.090 | 3.291 |
| | 0% | 50% | 60% | 70% | 80% | 90% | 95% | 98% | 99% | 99.8% | 99.9% |
| | Confidence Level | | | | | | | | | | |

Apply the previous formula:

$$\bar{x} \pm (t\text{critical value}) \cdot \frac{s}{\sqrt{n}} = [8, 10] \pm 2.11 \frac{[3, 4]}{\sqrt{18}}$$

$$= [8, 10] \pm \left[ \frac{2.11(3)}{\sqrt{18}}, \frac{2.11(4)}{\sqrt{18}} \right]$$

$$\eqsim [8, 10] \pm [1.492, 1.989].$$

Split the calculation into two possibilities:

$$[8, 10] + [1.492, 1.989] = [8 + 1.492, 10 + 1.989]$$
$$= [9.492, 11.989],$$

and

$$[8, 10] - [1.492, 1.989] = [8 - 1.989, 10 - 1.492]$$
$$= [6.011, 8.508].$$

Now we combine both results in a conservative way, and we get the neutrosophic t confidence interval for the population average of weight lifting: $[6.011, 11.989]$ kg.